\documentclass{article}
\usepackage{graphicx}
\usepackage{amsmath}
\usepackage{amsthm}
\usepackage{amssymb}
\usepackage{algorithm, algpseudocode}
\usepackage{xcolor}
\usepackage{hyperref}

\usepackage{tikz}
\usepackage{caption}
\usepackage{subcaption}
\usetikzlibrary{matrix}

\usepackage[normalem]{ulem}

\usepackage{booktabs}
\usepackage{environ}
\usepackage{multirow}
\usepackage{adjustbox}

\title{HJRNO: Hamilton-Jacobi Reachability with Neural Operators}

\author{
    \begin{minipage}[t]{0.45\textwidth}
        \centering
        \small
        \textbf{Yankai Li}\\
        School of Computing Science\\
        Simon Fraser University\\
        \texttt{yla890@sfu.ca}
        \vfill
    \end{minipage}
    \hfill
    \begin{minipage}[t]{0.45\textwidth}
        \centering
        \small
        \textbf{Mo Chen}\\
        School of Computing Science\\
        Simon Fraser University\\
        \texttt{mochen@cs.sfu.ca}
        \vfill
    \end{minipage}
}

\date{}

\begin{document}
\maketitle

\begingroup
\renewcommand\thefootnote{}\footnotetext{This work was supported by the Canada CIFAR AI Chairs and NSERC Discovery Grants Programs.}
\addtocounter{footnote}{-1}
\endgroup


\begin{abstract}

Ensuring the safety of autonomous systems under uncertainty is a critical challenge.
Hamilton-Jacobi reachability (HJR) analysis is a widely used method for guaranteeing safety under worst-case disturbances.
In this work, we propose HJRNO, a neural operator-based framework for solving backward reachable tubes (BRTs) efficiently and accurately.
By leveraging neural operators, HJRNO learns a mapping between value functions, enabling fast inference with strong generalization across different obstacle shapes and system configurations.
We demonstrate that HJRNO achieves low error on random obstacle scenarios and generalizes effectively across varying system dynamics.
These results suggest that HJRNO offers a promising foundation model approach for scalable, real-time safety analysis in autonomous systems.

\end{abstract}
\section{Introduction}

Autonomous systems are playing an increasingly significant role in daily life, drawing growing attention from both industry and academia. 
Despite their versatile capabilities and the substantial assistance they provide, ensuring their safety remains a critical concern. 
Many autonomous systems are exposed to unpredictable disturbances, such as varying weather conditions, which can impact their reliability and performance.
Hamilton-Jacobi (HJ) reachability analysis has emerged as a powerful tool for providing provable safety guarantees under worst-case disturbances, addressing both system configurations and control strategies \cite{bansal2017hamilton, chen2018hamilton}.

HJ reachability analysis achieves this by computing the backward reachable tube (BRT), which represents the set of states that the system must avoid to maintain safety. 
In addition to identifying unsafe regions, the BRT offers a quantitative measure of how close the current configuration is to unsafe states. 
Furthermore, the process of computing the BRT naturally yields an optimal control strategy that directs the system away from the unsafe region.

The computation of the BRT involves solving the Hamilton-Jacobi-Isaacs (HJI) partial differential equation (PDE) over the state space, typically discretized using a grid-based dynamic programming approach. 
This traditional method can produce highly accurate results when a high-resolution grid is used. 
However, the ``curse of dimensionality'' poses a significant challenge, as the number of grid points grows exponentially with the dimension of the system. 
Moreover, solving the HJI-PDE requires cross-dimensional computations, further compounding the computational burden.

Several research directions have been explored to accelerate the computation of the BRT. 
One approach involves computing an approximate BRT by decomposing the system into smaller subsystems, thereby avoiding cross-dimensional interactions \cite{chen2018decomposition, chen2017exact}. 
Another strategy simplifies the system dynamics, such as by linearizing nonlinear systems \cite{schurmann2017guaranteeing}. 
While these approximations can significantly reduce computational costs, they often lead to inaccuracies in the resulting BRT and typically require additional system-specific analysis to ensure safety.

Recently, deep learning-based approaches have been proposed to efficiently compute the BRT. The physics-informed machine learning framework \cite{bansal2021deepreach, singh2024exact, tayal2025physics} introduces neural PDE solvers that leverage neural networks to compute the BRT. 
However, these approaches typically learn the solution function for a single problem instance and must be retrained from scratch when the problem setting changes.
While \cite{borquez2023parameter} introduces a parameterized neural PDE solver, it only handles scalar hyperparameters rather than functions, limiting its applicability to more general settings where the input includes functions.

In this work, we propose solving the BRT using neural operators \cite{kovachki2023neural, li2020fourier, li2023fourier}, which learn mappings between infinite-dimensional function spaces. 
This approach offers two advantages: (1) the neural operator model needs to be trained only once and can subsequently generalize across a broad range of problem settings; (2) inference time is orders of magnitude faster than traditional BRT solvers.

Our main contributions are as follows:

\begin{enumerate}
    \item To the best of our knowledge, this is the first work to apply neural operators to Hamilton-Jacobi reachability (HJR) problems.
    \item By learning a neural operator, we achieve near-instantaneous inference (~$10^{-3}$ seconds), eliminating the need to solve the HJI-PDE for each different problem setting, such as varying obstacle shapes or system hyperparameters.
    \item We demonstrate that our method generalizes well across randomly generated obstacle shapes and varying system hyperparameters, indicating strong potential of neural operators for broad applicability in BRT problems.
\end{enumerate}

\section{Problem Setting}
\label{sec:prob}

We consider the dynamics of an autonomous system described by

\begin{subequations}
\label{eq:1}
\begin{align}
&\dot{x} = f(x, u, d), \\
& x(0) = x_0,\\
&x\in\mathbb{R}^n, \; u\in\mathcal{U},  \; d\in\mathcal{D}
\end{align}
\end{subequations}

\noindent where $x$ denotes the system state, $u$ the control input, and $d$ the external disturbance. 
The sets $\mathcal{U}$ and $\mathcal{D}$ denote the sets of measurable control and disturbance functions, respectively. We define the system trajectory starting from initial state $x$, subject to control $u(\cdot)$ and disturbance $d(\cdot)$ over the time interval $[t, \tau]$, as $\zeta(x, u(\cdot), d(\cdot), t, \tau)$. 
The unsafe set is defined as

\begin{equation}
\label{eq:6}
\mathcal{L} = \{x\;:\; l(x) \leq 0\},
\end{equation}

\noindent where the function $l(x): \mathbb{R}^n \rightarrow \mathbb{R}$ typically denotes the signed distance function. 
The backward reachable tube (BRT) is the set of all initial states from which, despite applying optimal control strategies aimed at avoiding $\mathcal{L}$, there exists a disturbance strategy that forces the system into the unsafe set within the time horizon $[t, \tau]$:

\begin{equation}
\label{eq:5}
\mathcal{R}(t) = \{x: \forall u(\cdot), \exists d(\cdot),
\exists \tau \in [t, T], \zeta(x, u(\cdot), d(\cdot), t, \tau) \in \mathcal{L} \}
\end{equation}

\subsection{Running examples}
\label{sec:examples}

\textbf{Air3D}.
This example models a two-vehicle collision avoidance scenario.
The system consists of two agents: an evader, which attempts to avoid capture, and a pursuer, which tries to intercept the evader.

The dynamics of the evader (agent $a$) and pursuer (agent $b$) are given by:

\begin{equation}
\begin{bmatrix}
\dot{x}_a\\
\dot{y}_a\\
\dot{\theta}_a
\end{bmatrix}
=
\begin{bmatrix}
v\cos(\theta_a)\\
v\sin(\theta_a)\\
u_a
\end{bmatrix},\quad
\begin{bmatrix}
\dot{x}_b\\
\dot{y}_b\\
\dot{\theta}_b
\end{bmatrix}
=
\begin{bmatrix}
v\cos(\theta_b)\\
v\sin(\theta_b)\\
u_b
\end{bmatrix}
\end{equation}

Here, $v$ is the constant speed shared by both agents, while $u_a$ and $u_b$ are the angular velocities of the evader and pursuer, respectively, with control bounds $u_a,\; u_b \in [-u_{\mathrm{max}},\, u_{\mathrm{max}}]$.

To simplify the problem, we can formulate the system dynamics in the evader's frame of reference, yielding a relative dynamics model:

\begin{equation}
\label{eq:7}
\begin{bmatrix}
\dot{x}_1\\
\dot{x}_2\\
\dot{x}_3
\end{bmatrix}
=
\begin{bmatrix}
-v_a + v_b \cos(x_3) + u_a x_2\\
v_b\sin(x_3) - u_a x_1\\
u_b - u_a
\end{bmatrix}
\end{equation}

In this relative system, $x_1$ and $x_2$ represent the position of the pursuer relative to the evader, and $x_3$ is the relative heading angle.

The unsafe set is defined as:

\begin{equation}
\mathcal{L} = \left\{ (x_1, x_2) \in \mathbb{R}^2 \;\middle|\; \|(x_1, x_2)\| \leq d \right\}
\end{equation}

\noindent which corresponds to the set of states where the evader and pursuer are within a distance $d$ of each other, i.e. a collision.

The backward reachable tube (BRT) computed from this unsafe set (\ref{eq:5}) represents the set of initial states from which the pursuer can guarantee a collision with the evader, despite the evader applying optimal control to avoid it.

\bigskip

\noindent\textbf{Dynamic Dubins car}. 
We model an autonomous vehicle using a generalized Dubins car model. 
Unlike the classical Dubins car, this formulation allows the vehicle's speed to vary and adopts a more realistic heading control mechanism based on curvature, which is standard for nonholonomic vehicles. 
Specifically, rather than directly controlling the turning rate, we apply a curvature control scaled by the velocity. 
In the following, we refer to this Dubins-like vehicle with acceleration and curvature control as the dynamic Dubins car. 
The system dynamics are given by:

\begin{subequations}
\label{eq:2}
\begin{align}
&\dot{x}_1 = v \cos(\theta) + d_1, \\
&\dot{x}_2 = v \sin(\theta) + d_2, \\
&\dot{v} = u_1, \\
&\dot{\theta} = v \,u_2
\end{align}
\end{subequations}

\noindent where the states are the position components $(x_1, x_2)$, the velocity $v$, and the heading angle $\theta$. The control inputs are the acceleration $u_1$ and the curvature $u_2$, while $d_1$ and $d_2$ represent disturbances acting on the $x_1$ and $x_2$ positions, respectively.

To define an unsafe set, consider the task of avoiding a tree obstacle. The unsafe set can be modeled as:

\begin{equation}
\mathcal{L} = \left\{ (x_1, x_2) \in \mathbb{R}^2 \;\middle|\; \|(x_1 - x_{\mathrm{tree}},\, x_2 - y_{\mathrm{tree}})\| \leq r \right\}
\end{equation}

\noindent where $(x_{\mathrm{tree}}, y_{\mathrm{tree}})$  denotes the center of the tree, and $r$ represents its effective radius. 
The BRT can then be computed by solving the HJI-PDE (\ref{eq:3}). 
Although this example assumes the obstacle (tree) has a circular shape, in the \nameref{sec:res} section, we demonstrate that our proposed method can handle obstacles of arbitrary shapes effectively.
\section{Background}
\label{sec:background}

\subsection{Hamilton-Jacobi Reachability}
To compute the BRT in~\eqref{eq:5}, we first define an objective function that measures the minimum value of $l$ (see Eq.~\eqref{eq:6} for its definition) along a trajectory:

\begin{equation}
J(t, x, u(\cdot), d(\cdot)) = \underset{\tau \in [t, T]}{\mathrm{min}} l (\zeta(t, \tau, x, u(\cdot), d(\cdot))) 
\end{equation}

The control strategy seeks to maximize this objective, effectively pushing the system away from unsafe states, while the disturbance seeks to minimize it. 
This leads to the definition of the value function:

\begin{subequations}
\begin{align}
V(t, x) &= \underset{d(\cdot)\in\mathbb{D}}{\mathrm{inf}}
\,\underset{u(\cdot)\in\mathbb{U}}{\mathrm{sup}} \; 
J\left(
t, x, u(\cdot), d(\cdot)
\right)
\\
V(T, x) &= l(x)
\end{align}
\end{subequations}

The value function $V(\cdot)$ is the viscosity solution of the Hamilton-Jacobi-Isaacs (HJI) partial differential equation (PDE) \cite{crandall1983viscosity, crandall1984some}:

\begin{equation}
\label{eq:3}
\min \left\{
\frac{\partial}{\partial t} V(t, x) + H(t, x),\; l(x) - V(t, x)
\right\} = 0
\end{equation}

\noindent where the Hamiltonian $H(t, x)$ is defined as

\begin{equation}
H(t, x) = \underset{u(\cdot) \in \mathbb{U}}{\mathrm{max}}\, \underset{d(\cdot)\in \mathbb{D}}{\mathrm{min}}\; \left(\frac{\partial}{\partial x} V(t, x)\right)^\top f(x, u(\cdot), d(\cdot))
\end{equation}

\subsection{Neural Operators}

Operator learning aims to learn mappings between input function and output function.
A naive approach would be to discretize these functions onto grids, and then learn a function approximator -- like conventional neural network architectures such as multilayer perceptrons (MLPs), convolutional neural networks (CNNs), or vision transformers (ViTs) -- that maps the discretized inputs to the discretized outputs.
However, this strategy inherently ties the model to the specific discretization used during training, limiting its ability to generalize across different resolutions or grid structures.

Neural operators are designed to directly learn mappings between function spaces without relying on fixed discretizations. 
As a result, neural operators can naturally adapt to new discretizations and varying resolutions without the need for retraining.

Neural operators are the only known models that offer a theoretical guarantee of both universal approximation and invariance to discretization \cite{kovachki2023neural}. 
In this context, universal approximation refers to the ability of neural operators to approximate any continuous operator between Banach spaces to arbitrary accuracy, while discretization invariance ensures the model generalizes across different resolutions without retraining.

Formally, the objective of neural operators is to learn a mapping between input and output functions defined over infinite-dimensional spaces.
Let $a \in \mathcal{A}$ and $s \in \mathcal{S}$ denote the input and output functions, respectively, where $\mathcal{A}$ and $\mathcal{S}$ are Banach spaces.
The true underlying operator is denoted by $\mathcal{M}^\ast$, such that

\begin{equation}
s(\cdot) = \mathcal{M}^\ast(a(\cdot))
\end{equation}

The goal is to approximate this operator using a parameterized model $\mathcal{M}_\theta:\; \mathcal{A} \rightarrow \mathcal{S}$, where $\theta \in \mathbb{R}^P$ denotes the learnable parameters. This approximation is obtained by minimizing the empirical risk over a dataset of $N$ input-output function pairs:

\begin{equation}
\label{eq:11}
\underset{\theta}{\mathrm{min}}\;
\frac{1}{N} \sum_{i=1}^N \left\|s_i - \mathcal{M}_\theta(a_i)\right\|^2_{\mathcal{S}}
\end{equation}

Several neural operator architectures have been proposed to approximate mappings between infinite-dimensional function spaces \cite{lu2019deeponet, li2020neural, hao2023gnot, cheng2024reference, azizzadenesheli2024neural, kovachki2023neural}.
In this work, we focus on two prominent approaches: Transformer-based Neural Operator (TNO) \cite{cao2021choose} and the Fourier Neural Operator (FNO) \cite{li2020fourier}.
TNO model complex spatial correlations via attention mechanisms, while FNOs use Fourier transforms to capture global structure in the spectral domain.

\subsubsection{Fourier Neural Operator}

The FNO constructs the mapping using a sequence of $L$ integral kernel operator blocks:

\begin{equation} \label{eq:4} 
s = \mathcal{M}_\theta(a) := \mathcal{P}_{\text{out}} \circ \text{Block}_L \circ \cdots \circ \text{Block}_1 \circ \mathcal{P}_{\text{in}} (a)
\end{equation}

\noindent where $a(\cdot)$ and $s(\cdot)$ are the input and output functions, $\mathcal{P}_{\text{in}}$ and $\mathcal{P}_{\text{out}}$ are shallow fully connected networks for projecting to and from higher-dimensional spaces.

Each block takes the form:

\begin{equation}
\text{Block}_i (f) (x) := \sigma \left( W_i \cdot f (x) + \mathcal{K}_{i} (f)(x) \right)
\end{equation}

\noindent where $\sigma$ is an activation function, $W_i$ is a learnable linear map, and $\mathcal{K}_{i}$ is an integral operator:

\begin{equation}
\mathcal{K}_i (f) (x) := \int_D \kappa_{\theta_i} (x - x') f (x') dx'
\end{equation}

By the convolution theorem \cite{bracewell1999fourier}, this operation is equivalent to:

\begin{equation}
\mathcal{K}_i (f) (x) = \mathcal{F}^{-1} \big(
\mathcal{F}(\kappa_{\theta_i}) \cdot \mathcal{F} (f)    
\big) (x)
\end{equation}

\noindent where $\mathcal{F}$ and $\mathcal{F}^{-1}$ denote the Fourier and inverse Fourier transforms, respectively.
FNO parameterizes $\mathcal{F}(\kappa_{\theta_i})$ using a learnable weight tensor $R_{\theta_i} \in \mathbb{R}^{d \times d}$, enabling global convolution in the spectral domain.

\subsubsection{Transformer-based Neural Operator}

The Transformer-based Neural Operator (TNO) construct mappings between function spaces using a sequence of attention based blocks.

We adopt the Galerkin Transformer architecture \cite{cao2021choose} as our implementation of the TNO in this work. 
The neural operator is defined as a composition of projection layers and attention-based blocks:

\begin{equation}
\label{eq:12}
s = \mathcal{M}_\theta(a) := \mathcal{P}_{\text{out}} \circ \text{Block}_L \circ \cdots \circ \text{Block}_1 \circ \mathcal{P}_{\text{in}} (a)
\end{equation}

Each block is constructed as:

\begin{equation}
\text{Block}_i(f) := f + \text{Attn}_i(f) + \mathcal{P}_i(f + \text{Attn}_i(f))
\end{equation}

\noindent where $\mathcal{P}_\text{in}$ $\mathcal{P}_\text{out}$, and $\mathcal{P}_i$ are shallow projection networks (MLPs), and $\text{Attn}_i$ denotes a self-attention operator at layer $i$.

The attention mechanism approximates an integral operator over the domain $D$ via:

\begin{equation}
\label{eq:10}
\text{Attn}(f)(x) \approx \sum_{i=1}^d \left(\int_D v(x') \cdot k_i(x') \, dx'\right) q_i(x) \approx \frac{1}{n} \sum_{i=1}^d \left(K_i^\top V\right) Q_{ij}
\end{equation}

Here

\begin{enumerate}
    \item $Q = X W_Q$, $K = XW_K$, and $V = X W_V$ are the query, key, and value matrices. Layer normalization is applied to $K$ and $V$.
    \item $X \in \mathbb{R}^{n \times d}$ is the input tensor, representing the function sampled at $n$ spatial points with $d$ dimensional features.
    \item $W_Q$, $W_K$, $W_V \in \mathbb{R}^{d \times d}$ are learnable projection weights.
    \item $K_i$ denotes the $i$th feature column of $K$.
    \item $Q_{ij}$ denotes the attention weight corresponding to the $i$th feature and the $j$th spatial location.
    In Eq.~\eqref{eq:10}, we interpret the spatial location indexed by $j$ as the point $x$ where the attention output is being evaluated.
\end{enumerate}

\section{Method}
\label{sec:method}

We use the Fourier Neural Operator (FNO) \cite{li2020fourier} and Transformer-based Neural Operator (TNO) \cite{cao2021choose} to learn a functional mapping for backward reachable tubes (BRTs).
We define the initial time as $t=0$ and the final time as $t=-T$.
We assume that the value function $V(t, x)$, which satisfies the HJI-PDE \eqref{eq:3}, converges as $T \rightarrow \infty$.
We denote this limiting solution as

\begin{equation}
V_\infty(x) := \lim_{T\rightarrow\infty}V(t=-T, x),
\end{equation}

\noindent which corresponds to the maximal BRT, and is most commonly used in practice for providing safety filters \cite{herbert2017fastrack, borquez2025dualguard}.

Our goal is to learn an operator $\mathcal{M}_\theta$ that maps the initial value function $V(t=0, x)$ to the solution $V_\infty(x)$:

\begin{subequations}
\label{eq:14}
\begin{align}
a(x) &:= V(t=0, \, x) \label{eq:8}\\
s(x) &:= V_\infty(x)
\end{align}
\end{subequations}

To construct the training dataset $\mathcal D = \left\{\left(V^{(i)}(t=0, x),\, V_\infty^{(i)}(x)\right)\right\}_{i=1}^N$, we solve the HJI-PDE in Eq.~\eqref{eq:3} for various problem instances using traditional dynamic programming solvers \cite{schmerling_hj_reachability}.
The Neural Operator model \eqref{eq:4} \eqref{eq:12} is then trained in a data-driven manner to approximate the mapping $\mathcal{M}_\theta: \;V(t=0, \, x) \mapsto V_\infty(x)$.

Both $V(t=0,\, x)$ and $V_\infty(x)$ are represented by uniformly sampling the value function over a bounded domain $D \subset \mathbb{R}^n$ in the state space of $x$. For each training function instance $i$, the function is represented as a set of values on a uniform grid:

\begin{equation}
\left\{V^{(i)}(t=0, x_1), V^{(i)}(t=0, x_2), \dots V^{(i)}(t=0, x_M)\right\}
\end{equation}

\noindent where $\{x_j\}^M_{j=1}$ are grid points in the domain $D$. The same sampling is applied to represent $V^{(i)}_{\infty}(x)$.

The model is trained by minimizing the mean squared error between predicted and ground truth value functions:

\begin{equation}
\underset{\theta}{\mathrm{min}}\;
\frac{1}{N} \sum_{i=1}^N \left\| V_\infty^{(i)}(x) - \mathcal{M}_\theta\left(V^{(i)}(t=0, x)\right) \right\|^2
\end{equation}

\subsection{Generalizing to Parametric Inputs}

Our method is able to handle scalar hyperparameters by embedding them as constant input functions over the state domain.
That is, for a given hyperparameter $h \in \mathbb{R}^m$ (e.g., the maximum acceleration limit in the dynamic Dubins car example from Section~\nameref{sec:examples}), we define an input function $\tilde{h}(x) = h$ for all $x \in D$. 
This allows our framework to incorporate parametric dependence using the same neural operator architecture.

In this setup, we train the model to learn the mapping:

\begin{subequations}
\label{eq:13}
\begin{align}
a(x) &:= V(t=0,\, x,\, h), \label{eq:9}\\
s(x) &:= V_\infty(x,\, h),
\end{align}
\end{subequations}

\noindent where the value function now depends on both the state $x$ and the hyperparameter $h$.
\section{Results} \label{sec:res}

We evaluate HJRNO on 6 distinct experimental setups, each designed to test generalization under different conditions such as random geometry and system dynamics.
In all experiments, the model takes the initial value function as input and predicts the value function at the infinite time horizon (see Eq.~\eqref{eq:14}).
This function typically varies only with the spatial coordinates $(x_1, x_2)$, while remaining approximately constant along other dimensions such as velocity or heading.

To improve efficiency, we treat the spatial coordinates $(x_1, x_2)$ as the primary input domain and encode the remaining state variables as constant hyperparameter $h \in \mathbb{R}^m$, following the formulation in Eq.~\eqref{eq:13}.
This approach significantly improves both performance and scalability for FNO and TNO models.
In the Air3D experiment, where the full state space includes position $(x, y)$ and heading $\theta$, using the full state led to notable degradation:
FNO's test error increased from 0.028 to 0.050, training time rose from 1 minute to 3 minutes, and model size grew from 38 MB to 454 MB;
Similarly, TNO's test error increased from 0.037 to 0.15, with training time rising from 4 to 6 minutes.

All experiments are conducted at a resolution of 50, with each model trained for 20 epochs. A summary of quantitative results across all setups is provided in Table~\ref{table:1} and Table~\ref{table:2}.
Our method demonstrates significant efficiency: 
the FNO checkpoint size is roughly equal to the memory required to store one single solution instance of the 4D Dubins car ($50 \times 50 \times 50 \times 50$ array, approx.~25MB), and the TNO model is even smaller.
Both FNO and TNO perform inference three orders of magnitude faster than traditional solvers.
This efficiency is particularly valuable in robotics applications, where onboard devices often have limited memory and require rapid, high-frequency updates as the environment evolves, such as when obstacles move or change shape.

\begin{table}[htbp]
\centering
\caption{Performance metrics. 
\textbf{Train Err} and \textbf{Test Err} report relative $L_2$ errors on the training and test datasets, respectively. 
\textbf{Dataset Time} refers to the time required to generate both training and testing datasets using traditional solvers.
\textbf{Train Time} denotes the total training duration for 20 epochs. 
\textbf{Params} indicates the number of trainable parameters in each model.}
\label{table:1}
\renewcommand{\arraystretch}{1.2}
\begin{adjustbox}{max width=\textwidth}
\begin{tabular}{l l|ccccc}
\toprule
\textbf{Method} & \textbf{Experiment} &
Train Err & Test Err & Dataset Time & Train Time & Params \\
\midrule
\multirow{6}{*}{\rule{0pt}{12pt}FNO}
& Air3D & 0.017 & 0.028 & 50 min& 1 min & 4.83M \\
& Single Obstacle & 0.0019 & 0.0092 & 3 min & 17 min & 4.83M \\
& Two Obstacles & 0.0020 & 0.014 & 3 min & 17 min & 4.83M \\
& Indoor Environment & 0.0035 & 0.029 & 20 min & 40 min & 4.83M \\
& Velocity-Dependent & 0.0021 & 0.035 & 11 min & 40 min & 4.83M \\
& Parametric Inputs & 0.0007 & 0.016 & 3 min & 20 min & 4.83M \\
\midrule
\multirow{6}{*}{\rule{0pt}{12pt}TNO}
& Air3D & 0.034 & 0.037 & 50 min& 4 min & 0.8M\\
& Single Obstacle & 0.0014 & 0.0020 & 3 min& 80 min & 0.8M \\
& Two Obstacles & 0.0024 & 0.0063 &3 min & 80 min & 0.8M \\
& Indoor Environment & 0.0062 & 0.023 &20 min & 190 min & 0.8M \\
& Velocity-Dependent & 0.0025 & 0.044 & 11 min& 187 min & 0.8M \\
& Parametric Inputs & 0.0004 & 0.013 & 3 min& 95 min & 0.8M \\
\bottomrule
\end{tabular}
\end{adjustbox}
\end{table}

\begin{table}[htbp]
\centering
\caption{Resource usage metrics.
\textbf{VRAM (Train)} denotes the peak GPU memory usage during training.
\textbf{Data Size} refers to the storage size of the training dataset.
\textbf{Chkpt Size} is the size of the saved model checkpoint.
\textbf{Solve Time} is the time required to compute a single solution using traditional HJR solvers.
\textbf{Inference Time} is the time required for the trained model to produce a prediction.
}
\label{table:2}
\renewcommand{\arraystretch}{1.2}
\begin{adjustbox}{max width=\textwidth}
\begin{tabular}{l l|ccccc}
\toprule
\textbf{Method} & \textbf{Experiment} &
VRAM (Train) & Data Size & Chkpt Size &
Solve Time & Inference Time \\
\midrule
\multirow{6}{*}{\rule{0pt}{12pt}FNO}
& Air3D & 0.5 GB & 49 MB & 38 MB & 0.05 s & 0.001 s \\
& Single Obstacle & 0.5 GB & 2.4 GB & 38 MB & 0.9 s & 0.003 s \\
& Two Obstacles & 0.5 GB& 2.4 GB& 38 MB & 1.0 s & 0.002 s \\
& Indoor Environment & 0.5 GB & 7.3 GB & 38 MB & 2.6 s & 0.003 s \\
& Velocity-Dependent & 0.5 GB & 7.3 GB & 38 MB & 1.0 s & 0.002 s \\
& Parametric Inputs & 0.5 GB& 2.4 GB & 38 MB & 0.9 s & 0.002 s \\
\midrule
\multirow{6}{*}{\rule{0pt}{12pt}TNO}
& Air3D & 1.8 GB & 49 MB & 3 MB& 0.05 s & 0.003s \\
& Single Obstacle & 1.8 GB & 2.4 GB& 3 MB & 0.9 s & 0.004 s \\
& Two Obstacles & 1.8 GB & 2.4 GB& 3 MB & 1.0 s & 0.003 s \\
& Indoor Environment & 1.8 GB & 7.3 GB & 3 MB & 2.6 s & 0.005 s\\
& Velocity-Dependent & 1.8 GB &7.3 GB & 3 MB & 1.0 s & 0.004 s \\
& Parametric Inputs & 1.8 GB & 2.4 GB& 3 MB & 0.9 s & 0.005 s \\
\bottomrule
\end{tabular}
\end{adjustbox}
\end{table}

\subsection{Air3D Collision Avoidance}

The Air3D problem setup is detailed in Section~\nameref{sec:examples}.
In this experiment, we test HJRNO's ability to generalize over varying agent geometries.
Specifically, we randomly generate arbitrary smooth shapes for both the evader and the pursuer, as visualized in Fig.~\ref{fig:1}.

Each shape is created by sampling random radii at evenly spaced angles in polar coordinates, applying a convex hull to ensure a realistic boundary, and then smoothing the resulting polygon using cubic B-spline interpolation.

\begin{figure}[h]
    \centering
    \begin{tikzpicture}
        \matrix[matrix of nodes,
                nodes={inner sep=0, outer sep=0},
                column sep=5mm, row sep=5mm] {
            \node{}; &
            \node{\textbf{Ground Truth}}; &
            \node{\textbf{Prediction}}; &
            \node{\textbf{Error Map}}; \\
            \node{
                \begin{tikzpicture}
                    \node[rotate=90, anchor=center]{\textbf{FNO}};
                \end{tikzpicture}
            }; &
            \node{\includegraphics[width=3cm]{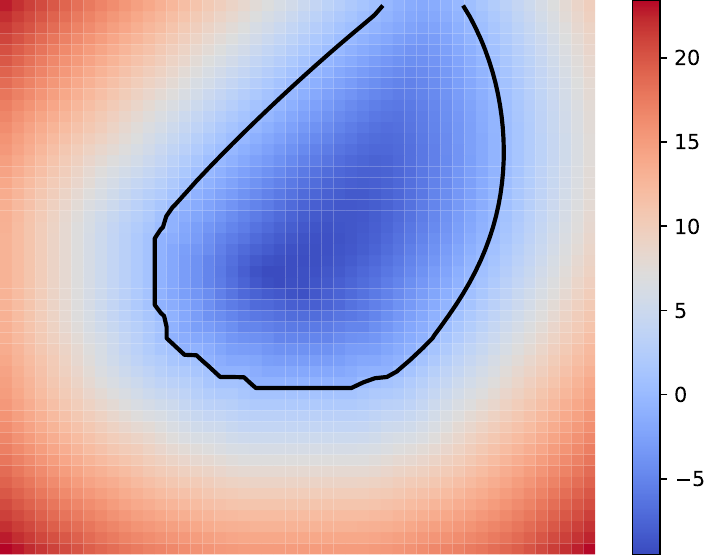}}; &
            \node{\includegraphics[width=3cm]{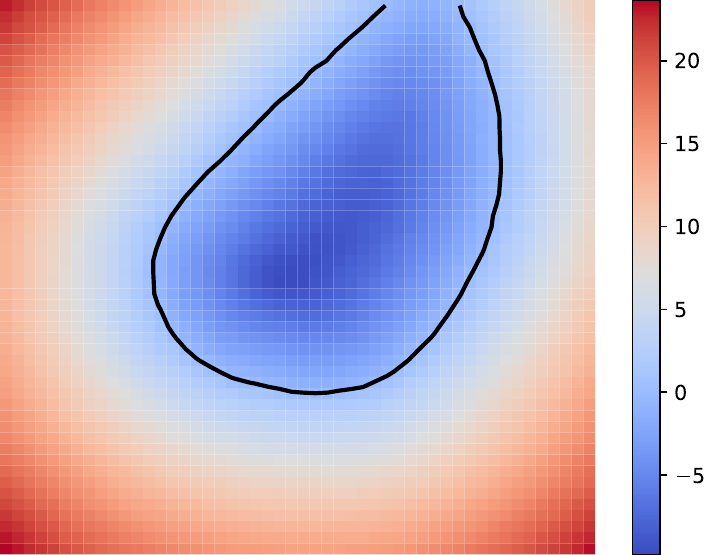}}; &
            \node{\includegraphics[width=3cm]{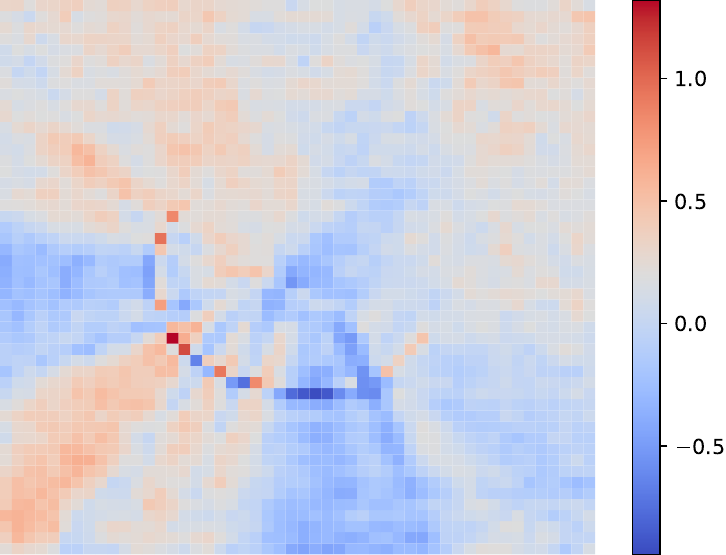}}; \\
            \node{
                \begin{tikzpicture}
                    \node[rotate=90, anchor=center]{\textbf{TNO}};
                \end{tikzpicture}
            }; &
            \node{\includegraphics[width=3cm]{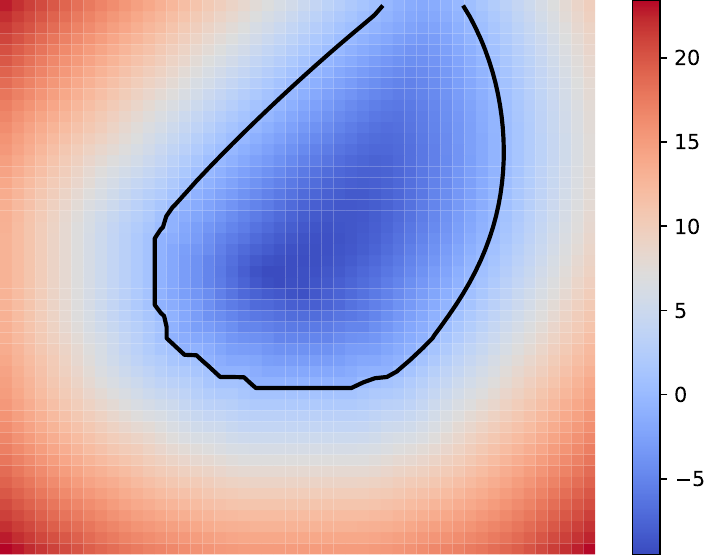}}; &
            \node{\includegraphics[width=3cm]{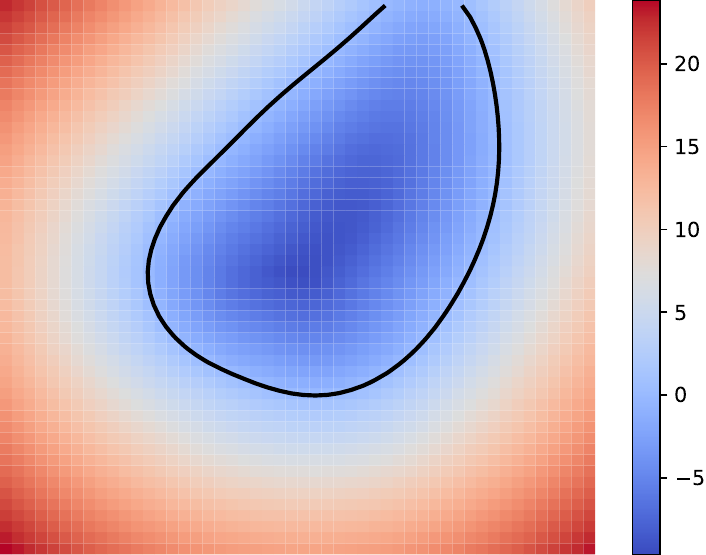}}; &
            \node{\includegraphics[width=3cm]{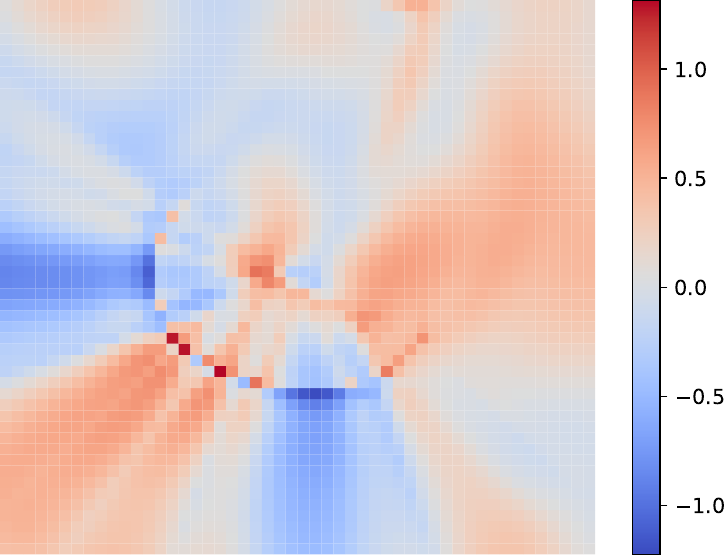}}; \\
        };
    \end{tikzpicture}
    \caption{Air3D}
    \label{fig:8}
\end{figure}

Using these shapes, we generate 100 random pairs of evader and pursuer geometries.
The training dataset consists of the corresponding $(V(t=0, x), V_\infty(x))$ pairs.
The model is trained on these 100 samples and evaluated on 50 additional, unseen pairs.
Examples of results are shown in Fig.~\ref{fig:8}.

\subsection{Dynamic Dubins car}

All five experiments in this section are based on the dynamics Dubins car model described in the \nameref{sec:examples} section.

\begin{figure}[h]
    \centering
    \begin{subfigure}[b]{0.7\textwidth}
        \centering
        \includegraphics[width=\linewidth]{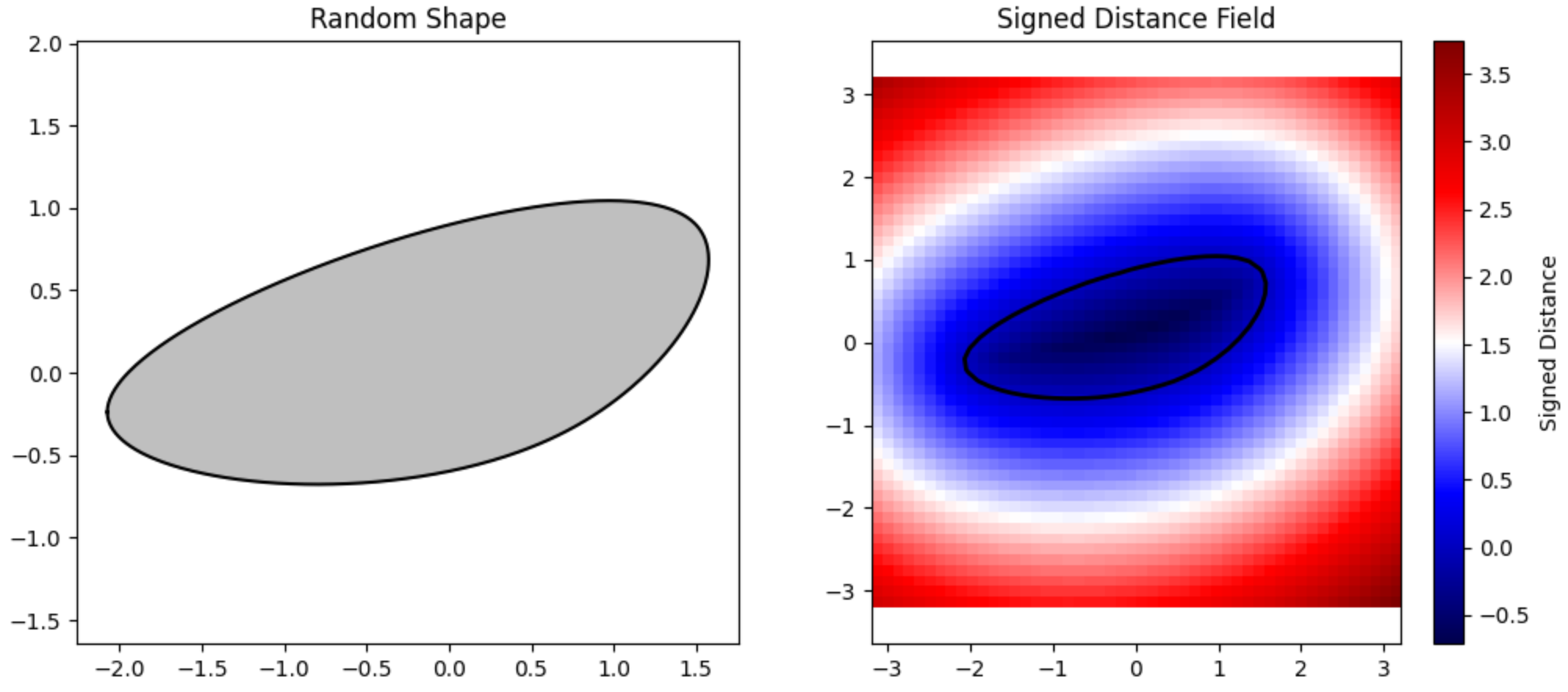}
    \end{subfigure}
    
    \vspace{5mm}
    
    \begin{subfigure}[b]{0.7\textwidth}
        \centering
        \includegraphics[width=\linewidth]{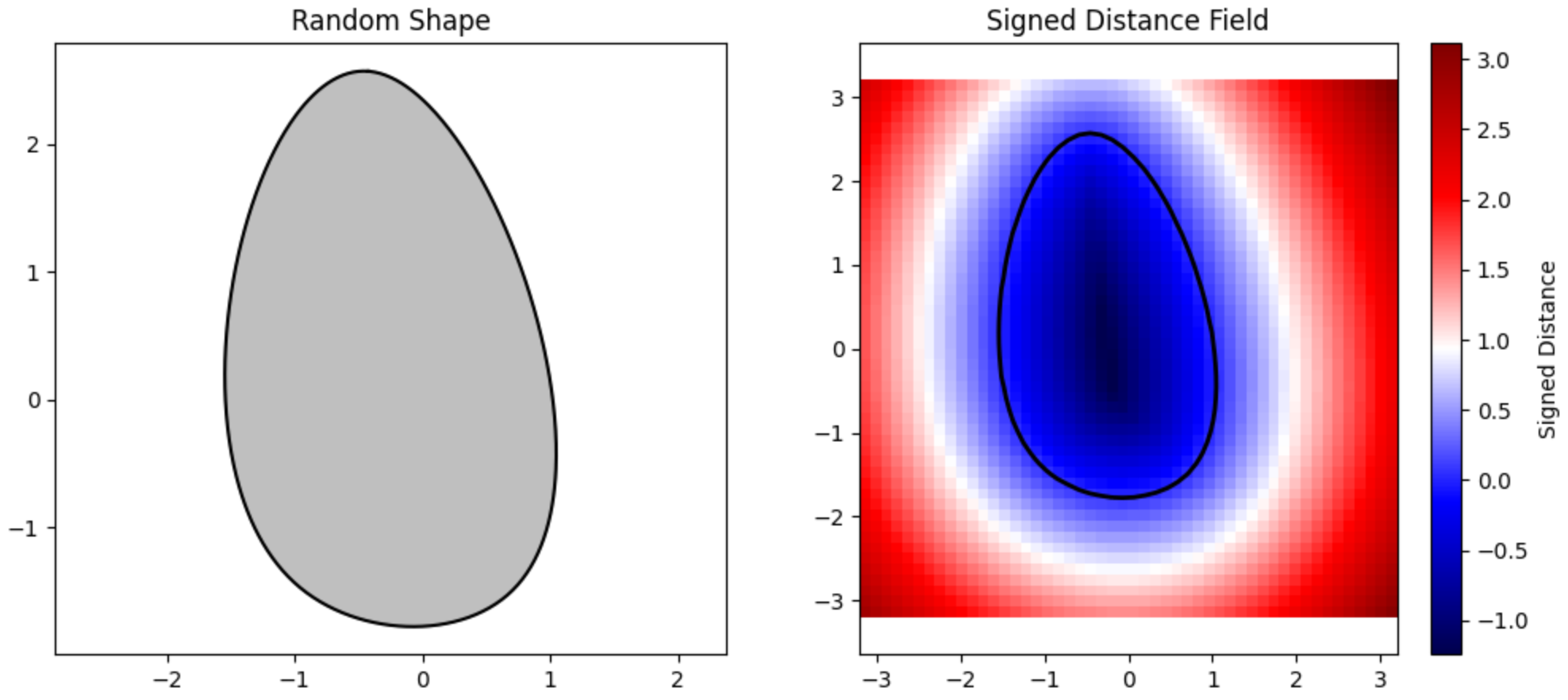}
    \end{subfigure}
    
    \vspace{5mm}
    
    \begin{subfigure}[b]{0.7\textwidth}
        \centering
        \includegraphics[width=\linewidth]{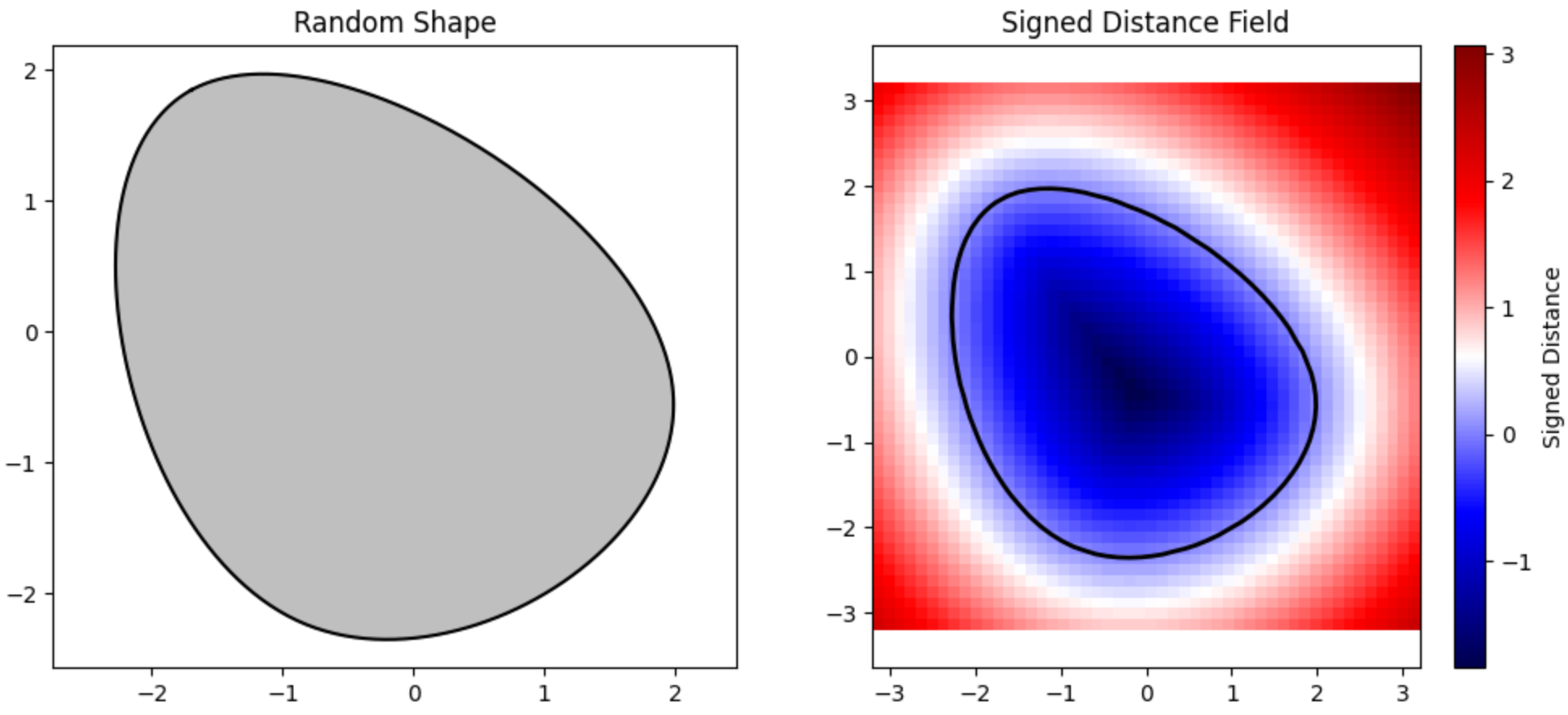}
    \end{subfigure}
    
    \caption{Random obstacle shapes}
    \label{fig:1}
\end{figure}

\begin{figure}[h]
    \centering
    \begin{tikzpicture}
        \matrix[matrix of nodes,
                nodes={inner sep=0, outer sep=0},
                column sep=5mm, row sep=5mm] {
            \node{}; &
            \node{\textbf{Ground Truth}}; &
            \node{\textbf{Prediction}}; &
            \node{\textbf{Error Map}}; \\
            \node{
                \begin{tikzpicture}
                    \node[rotate=90, anchor=center]{\textbf{FNO}};
                \end{tikzpicture}
            }; &
            \node{\includegraphics[width=3cm]{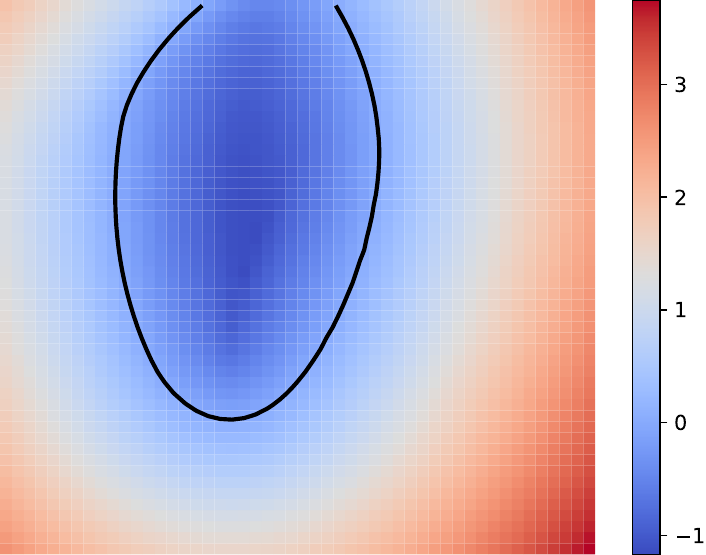}}; &
            \node{\includegraphics[width=3cm]{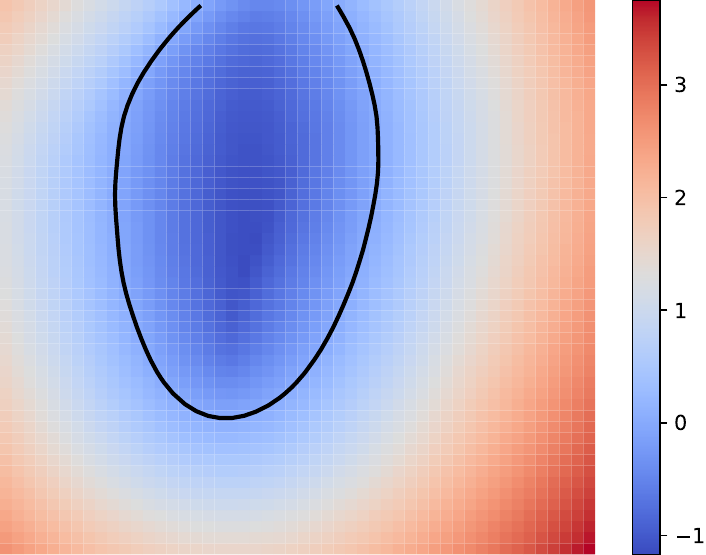}}; &
            \node{\includegraphics[width=3cm]{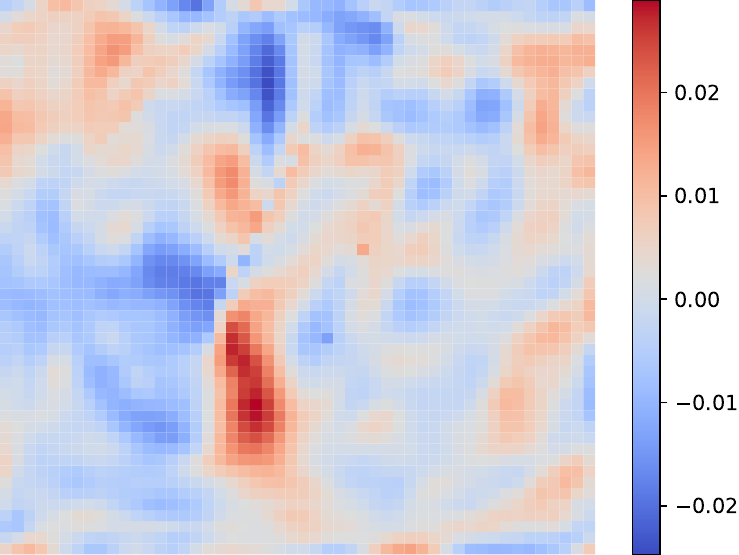}}; \\
            \node{
                \begin{tikzpicture}
                    \node[rotate=90, anchor=center]{\textbf{TNO}};
                \end{tikzpicture}
            }; &
            \node{\includegraphics[width=3cm]{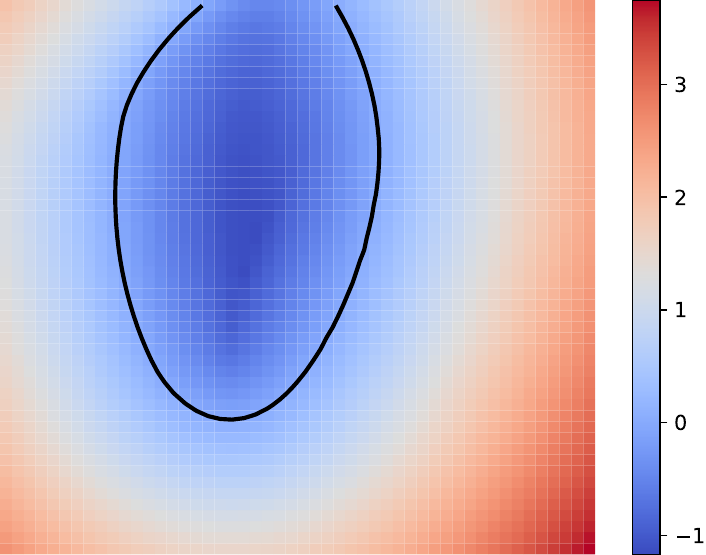}}; &
            \node{\includegraphics[width=3cm]{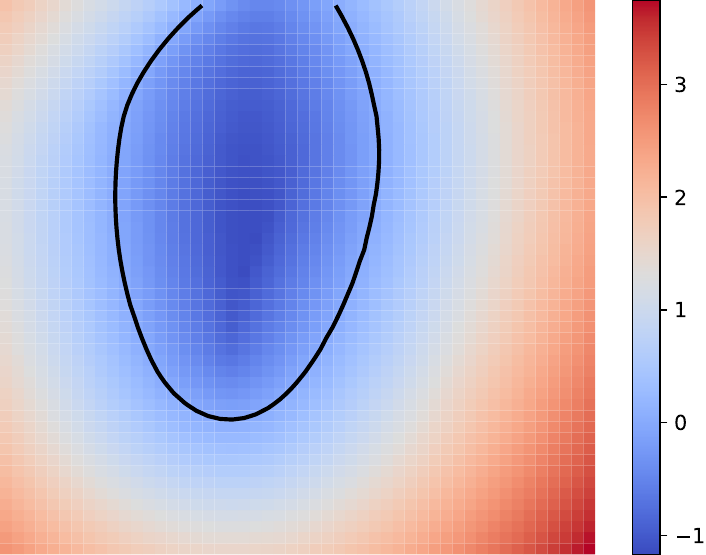}}; &
            \node{\includegraphics[width=3cm]{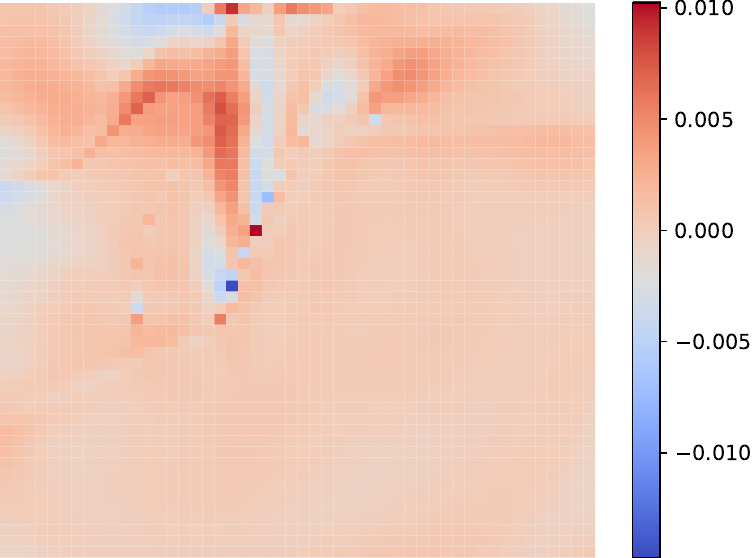}}; \\
        };
    \end{tikzpicture}
    \caption{Single Obstacle}
    \label{fig:2}
\end{figure}

\subparagraph{Single Obstacle.}
We train on 100 randomly generated smooth obstacles and test on 50.
Obstacles are created using the same random shape generation method as in the Air3D experiment (see Fig.\ref{fig:1}).
Examples of results are shown in Fig.\ref{fig:2}.

\begin{figure}[h]
    \centering
    \includegraphics[width=0.8\textwidth]{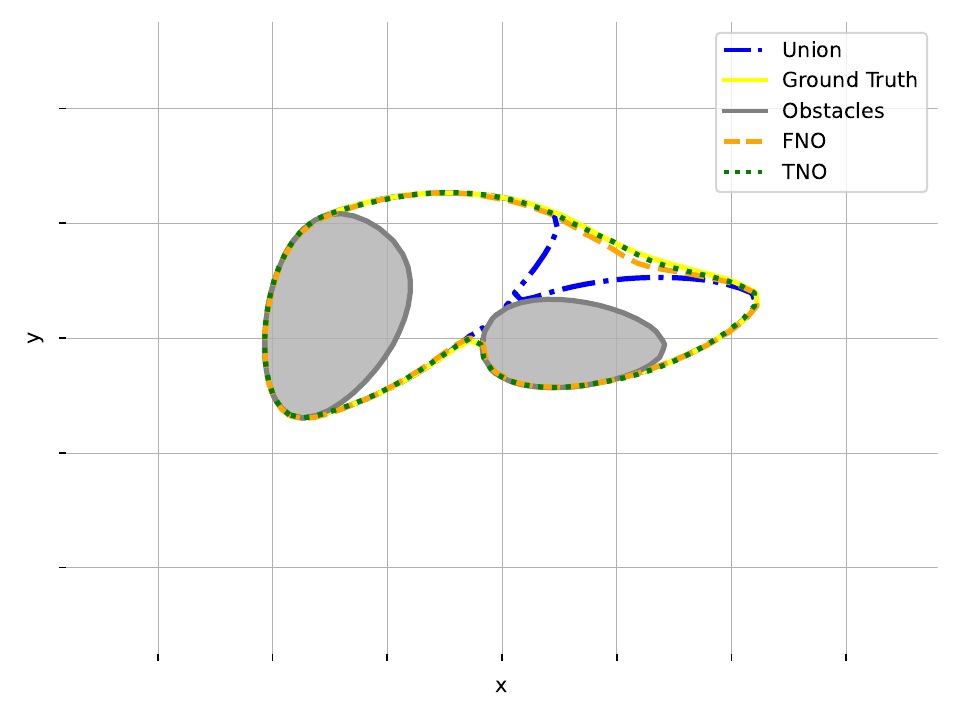}
    \caption{}
    \label{fig:11}
\end{figure}

\begin{figure}[h]
    \centering
    \begin{subfigure}[b]{0.7\textwidth}
        \centering
        \includegraphics[width=\linewidth]{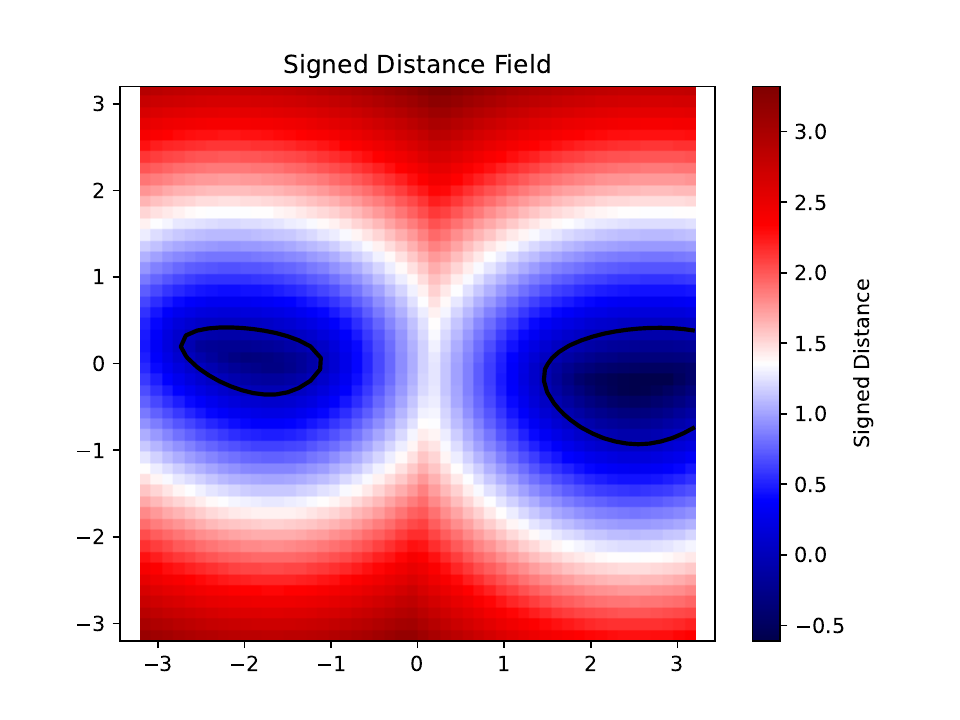}
    \end{subfigure}
    
    \vspace{5mm}
    
    \begin{subfigure}[b]{0.7\textwidth}
        \centering
        \includegraphics[width=\linewidth]{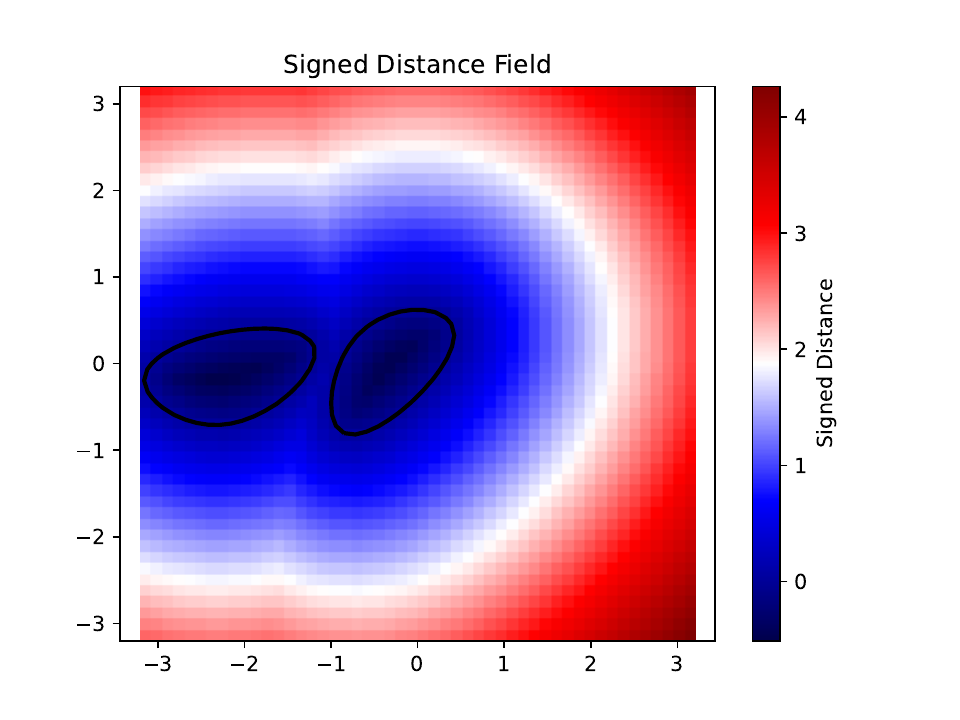}
    \end{subfigure}
    
    \vspace{5mm}
    
    \begin{subfigure}[b]{0.7\textwidth}
        \centering
        \includegraphics[width=\linewidth]{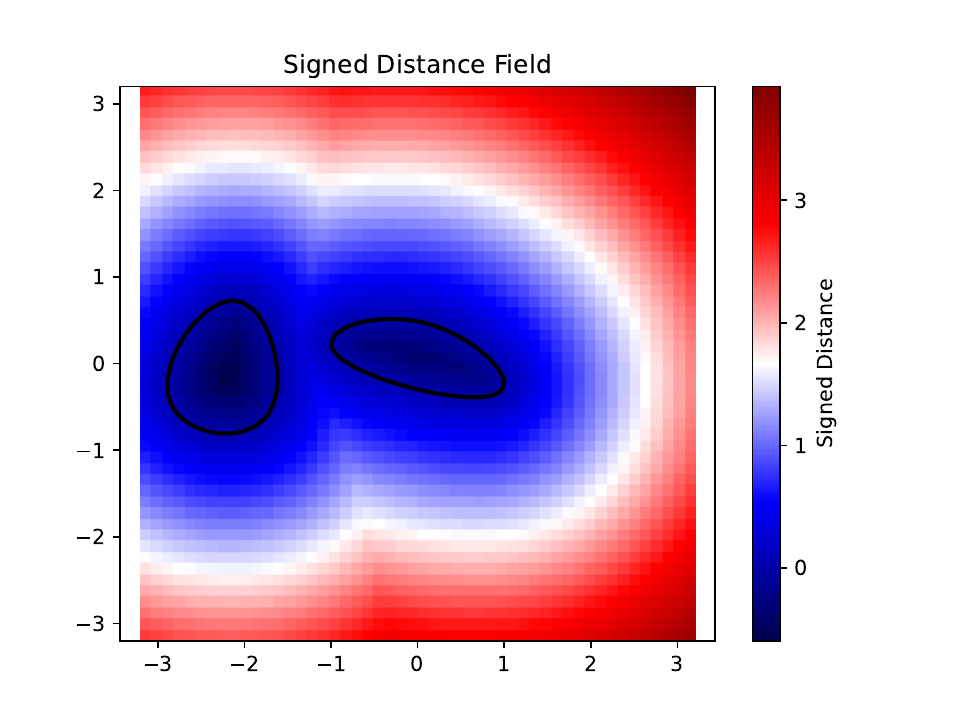}
    \end{subfigure}
    
    \caption{Two Obstacles}
    \label{fig:6}
\end{figure}

\subparagraph{Two Obstacles.}
This experiment introduces a second obstacle with a randomly sampled separation distance from the first.
A visualization is shown in Fig.~\ref{fig:6}.
Note that simply taking the union of the solutions for each obstacle individually does not yield the correct solution for the combined two-obstacle scenario, as shown by Fig.~\ref{fig:11}.
We generate 100 training and 50 testing samples of $(V(t=0, x), V_\infty(x))$ pairs.

\begin{figure}[h]
    \centering
    \begin{tikzpicture}
        \matrix[matrix of nodes,
                nodes={inner sep=0, outer sep=0},
                column sep=5mm, row sep=5mm] {
            \node{}; &
            \node{\textbf{Trained Resolution}}; &
            \node{\textbf{3x Larger Resolution}}; \\
            \node{
                \begin{tikzpicture}
                    \node[rotate=90, anchor=center]{\textbf{FNO}};
                \end{tikzpicture}
            }; &
            \node{\includegraphics[width=7cm]{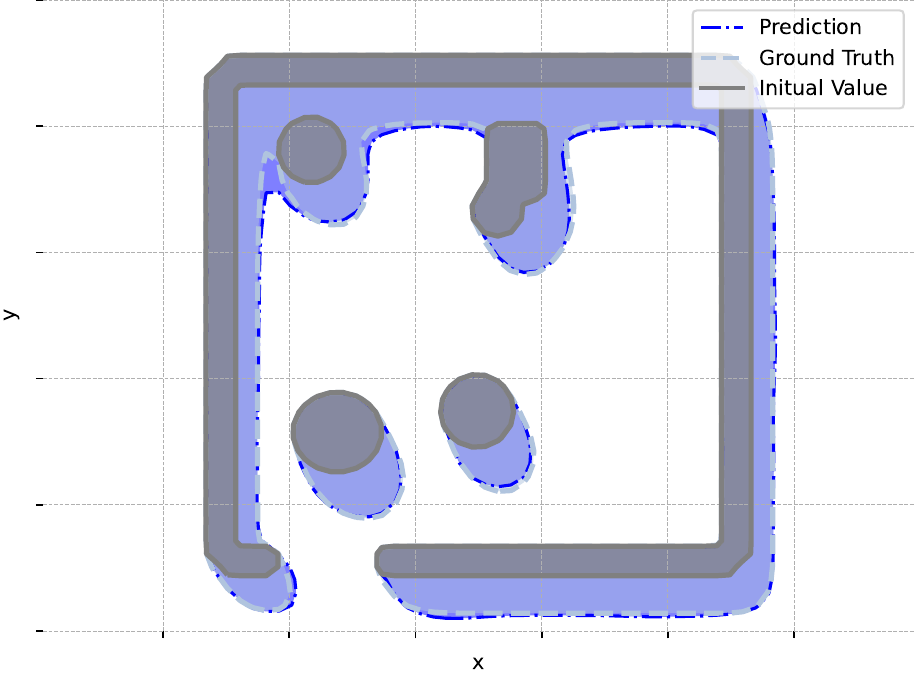}}; &
            \node{\includegraphics[width=7cm]{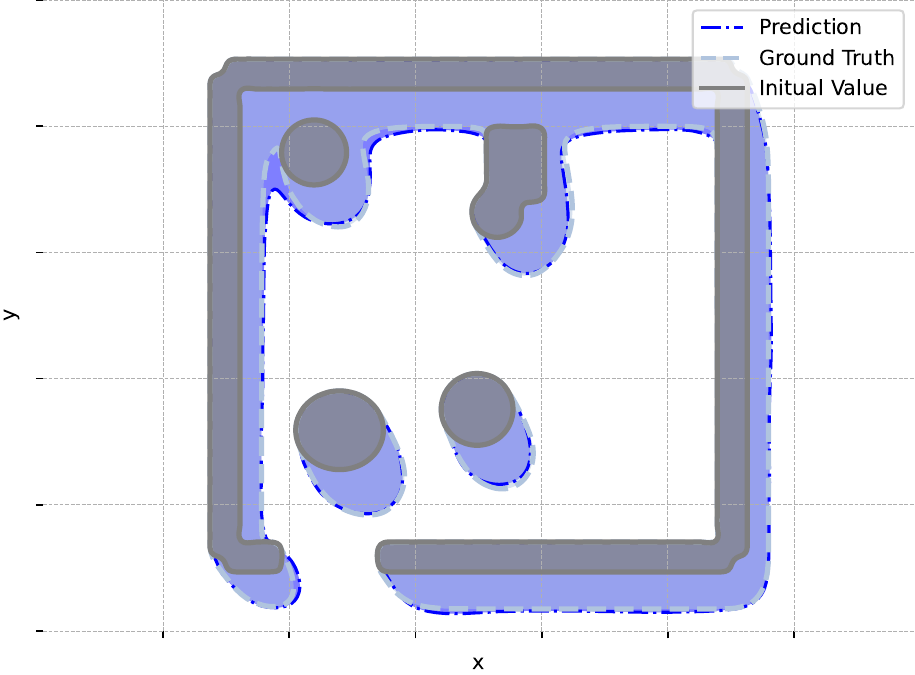}};\\
            \node{
                \begin{tikzpicture}
                    \node[rotate=90, anchor=center]{\textbf{TNO}};
                \end{tikzpicture}
            }; &
            \node{\includegraphics[width=7cm]{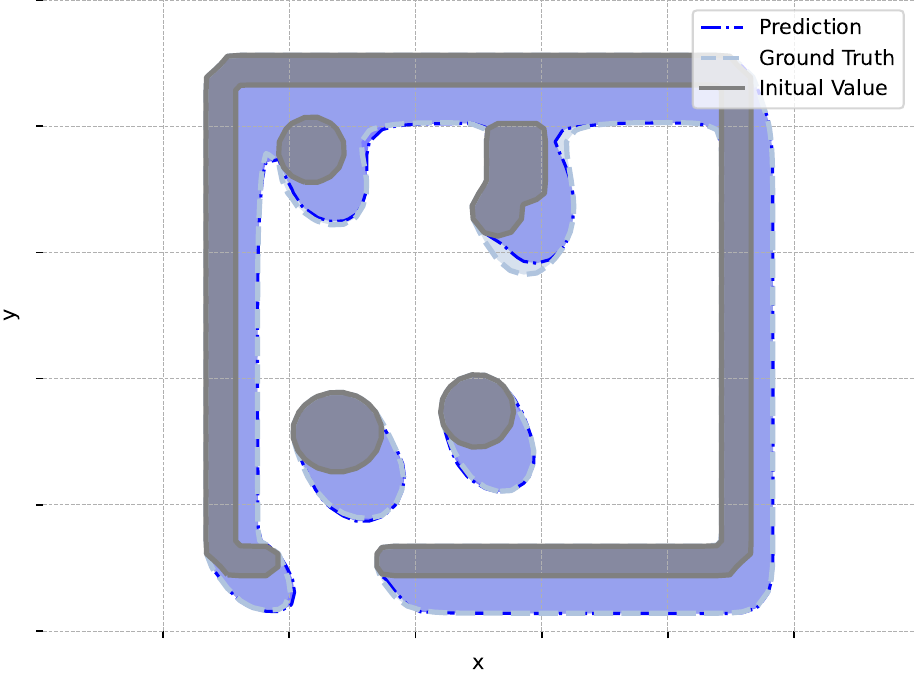}}; &
            \node{\includegraphics[width=7cm]{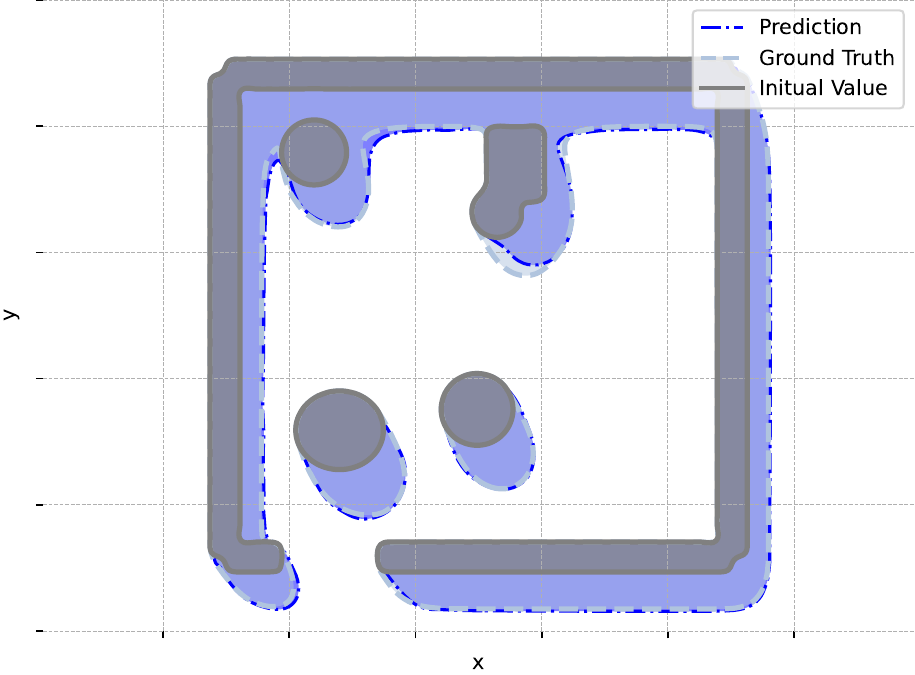}}; \\
            \node{
                \begin{tikzpicture}
                    \node[rotate=90, anchor=center]{\textbf{CNN}};
                \end{tikzpicture}
            }; &
            \node{\includegraphics[width=7cm]{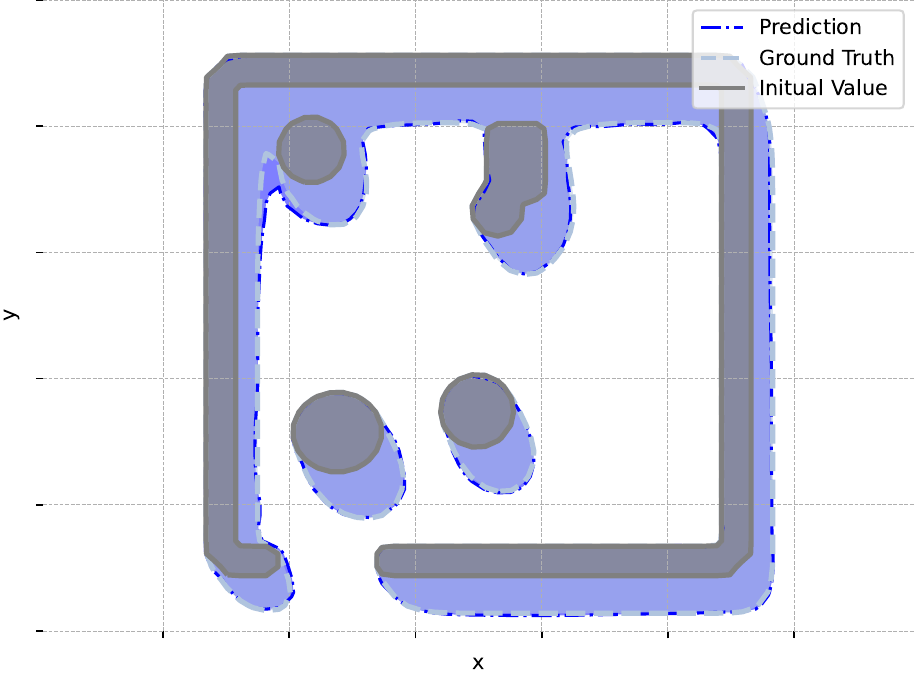}}; &
            \node{\includegraphics[width=7cm]{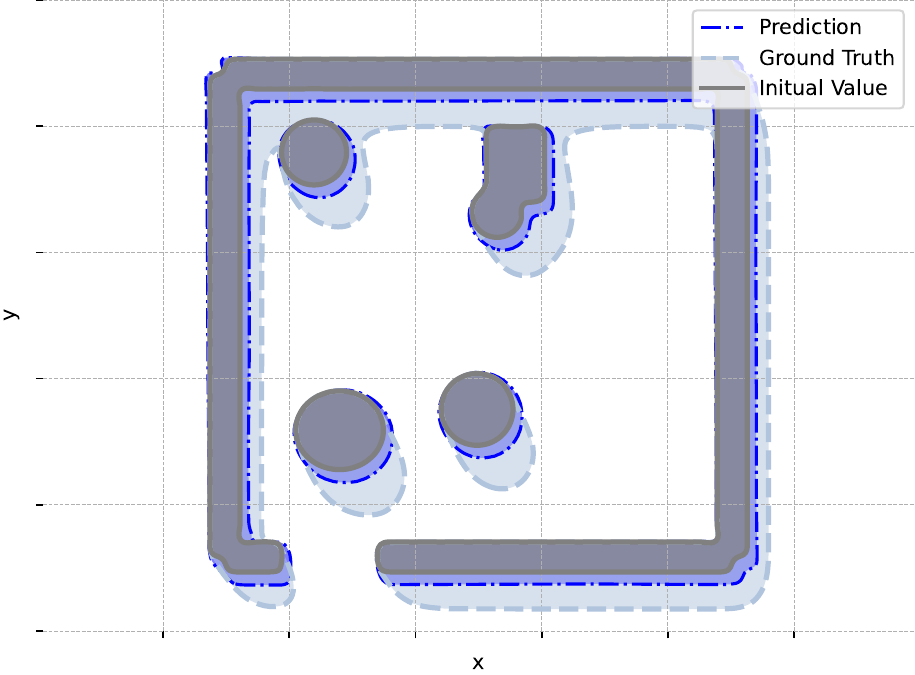}}; \\
        };
    \end{tikzpicture}
    \caption{Zero-shot super-resolution: Unlike conventional neural networks such as CNNs, our neural operators (FNO and TNO) learn mappings between function spaces, enabling zero-shot super-resolution.
    This allows models trained on low-resolution data to generalize directly to higher resolutions without retraining.}
    \label{fig:13}
\end{figure}

\begin{figure}[h]
    \centering
    \begin{tikzpicture}
        \matrix[matrix of nodes,
                nodes={inner sep=0, outer sep=0},
                column sep=5mm, row sep=5mm] {
            \node{}; &
            \node{\textbf{FNO}}; &
            \node{\textbf{TNO}}; \\
            \node{
                \begin{tikzpicture}
                    \node[rotate=90, anchor=center]{\textbf{Comparison}};
                \end{tikzpicture}
            }; &
            \node{\includegraphics[width=7cm]{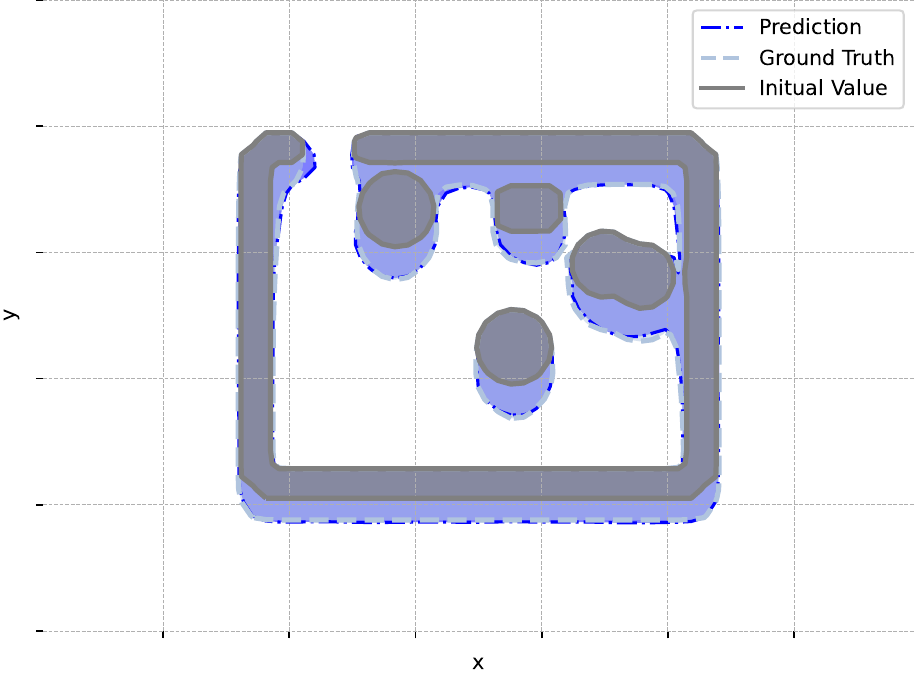}}; &
            \node{\includegraphics[width=7cm]{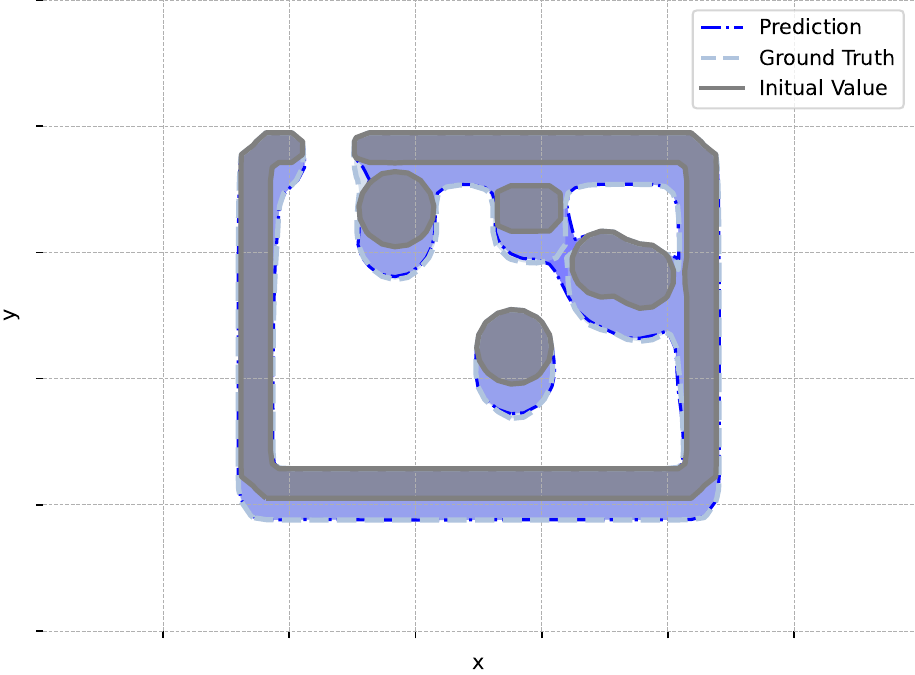}};\\
            \node{
                \begin{tikzpicture}
                    \node[rotate=90, anchor=center]{\textbf{Intial Values}};
                \end{tikzpicture}
            }; &
            \node{\includegraphics[width=6cm]{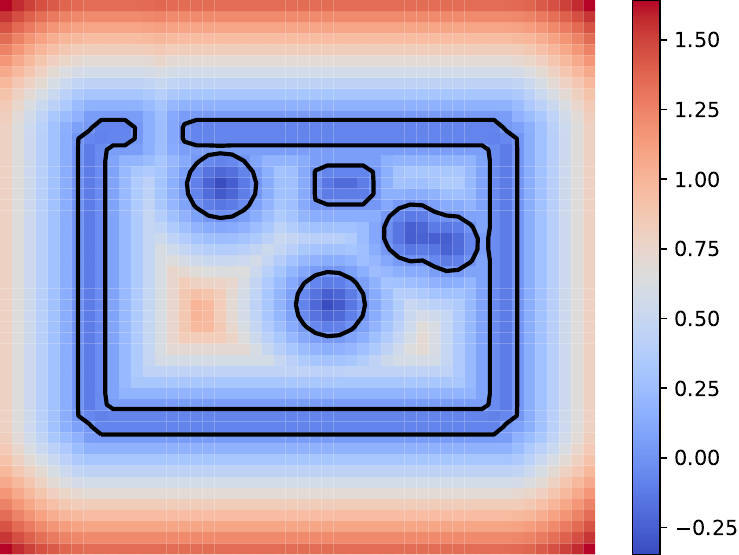}}; &
            \node{\includegraphics[width=6cm]{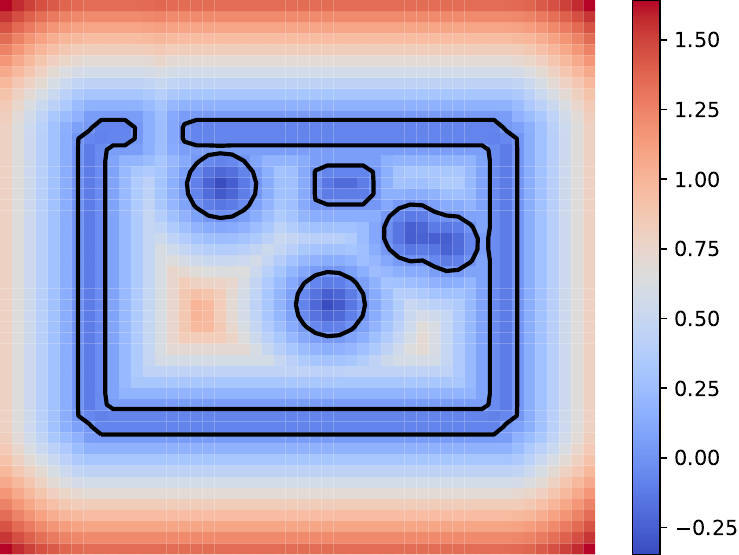}}; \\
            \node{
                \begin{tikzpicture}
                    \node[rotate=90, anchor=center]{\textbf{Ground Truth}};
                \end{tikzpicture}
            }; &
            \node{\includegraphics[width=6cm]{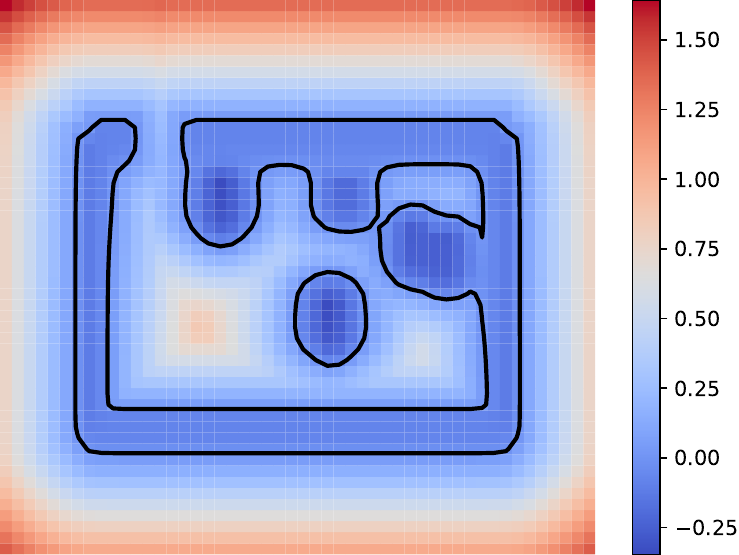}}; &
            \node{\includegraphics[width=6cm]{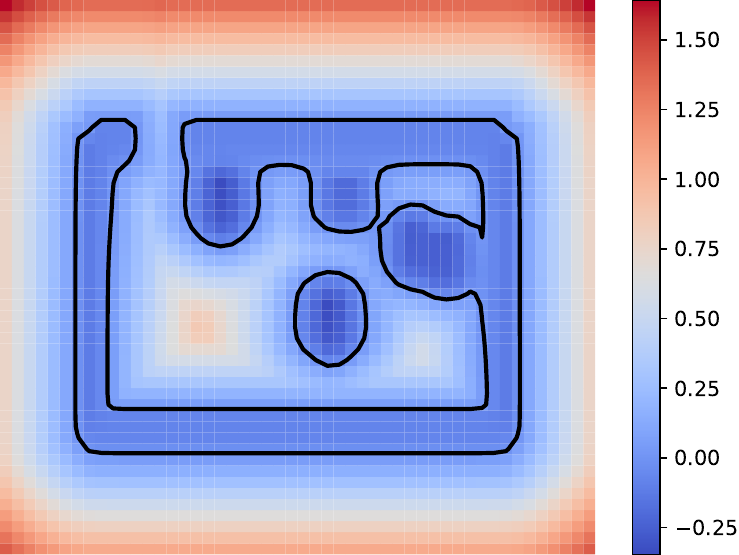}}; \\
            \node{
                \begin{tikzpicture}
                    \node[rotate=90, anchor=center]{\textbf{Prediction}};
                \end{tikzpicture}
            }; &
            \node{\includegraphics[width=6cm]{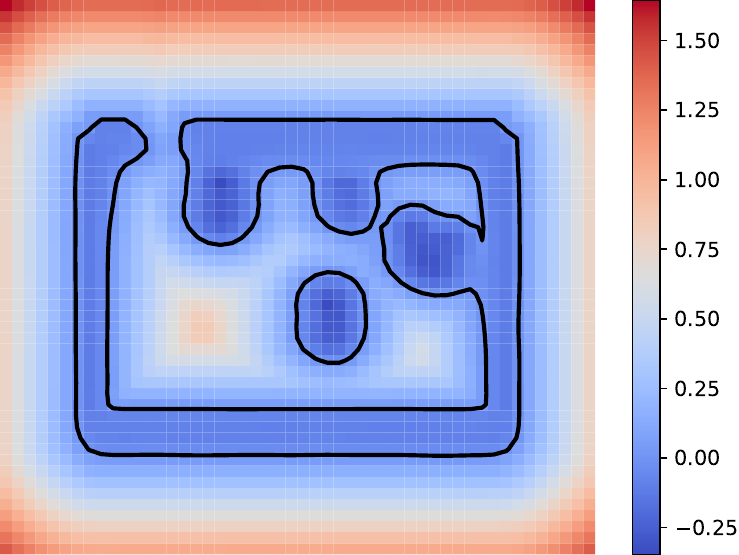}}; &
            \node{\includegraphics[width=6cm]{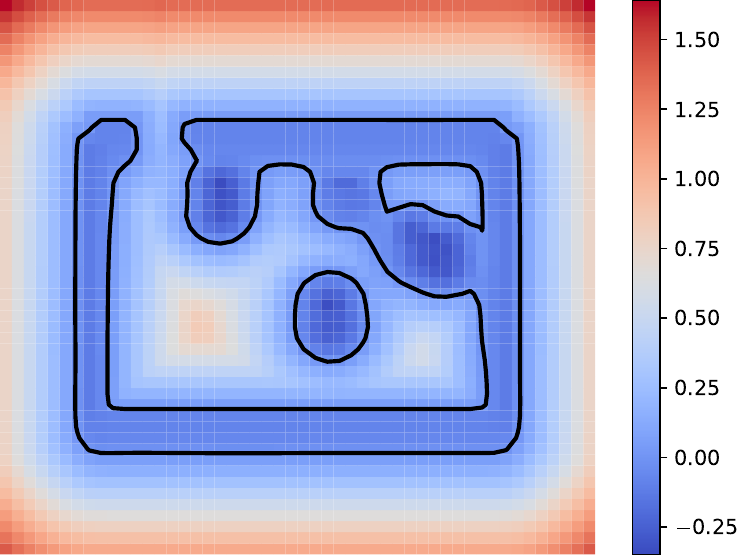}}; \\
        };
    \end{tikzpicture}
    \caption{Indoor Environment}
    \label{fig:12}
\end{figure}

\subparagraph{Indoor Environment.}
We simulate indoor spaces featuring walls, doors, and interior clutter.
Room layouts are generated with randomized door placements and walls, and obstacles inside the room include randomly sized boxes, circles, ellipses.
We generate 300 training and 50 testing $(V(t=0, x), V_\infty(x))$ pairs. Examples of results are shown in Fig.~\ref{fig:12}. Fig.~\ref{fig:13} illustrates that neural operators can perform zero-shot super-resolution, whereas conventional networks (e.g., CNNs) fail to generalize across resolutions.

\begin{figure}[h]
    \centering
    \begin{subfigure}[b]{0.7\textwidth}
        \centering
        \includegraphics[width=\linewidth]{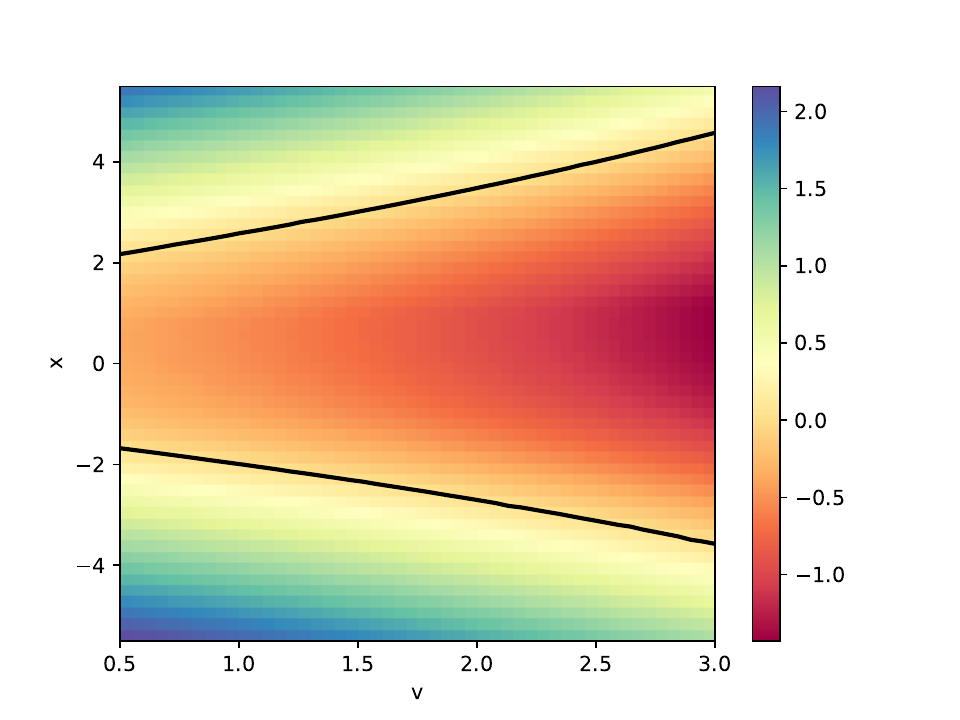}
    \end{subfigure}
    
    \vspace{5mm}
    
    \begin{subfigure}[b]{0.7\textwidth}
        \centering
        \includegraphics[width=\linewidth]{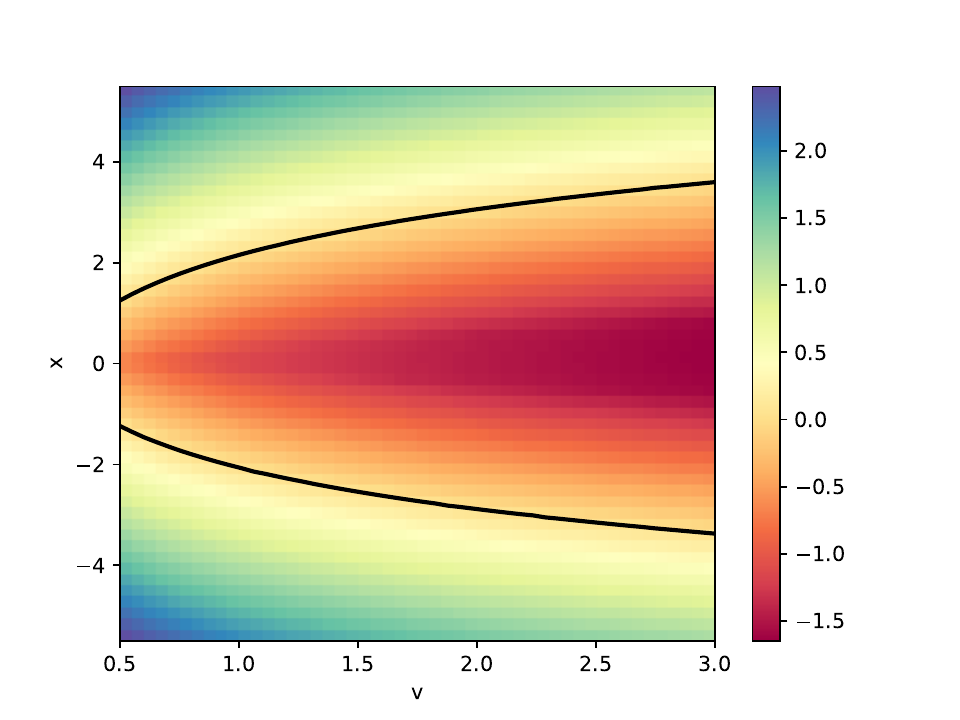}
    \end{subfigure}
    
    \vspace{5mm}
    
    \begin{subfigure}[b]{0.7\textwidth}
        \centering
        \includegraphics[width=\linewidth]{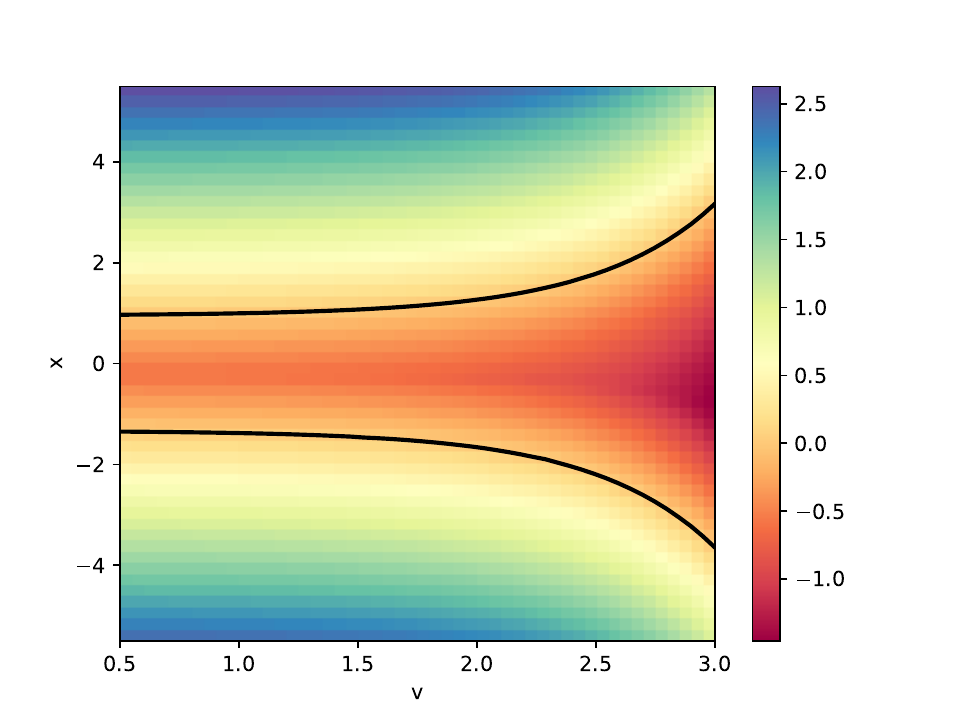}
    \end{subfigure}
    
    \caption{Velocity-Dependent}
    \label{fig:7}
\end{figure}

\begin{figure}[h]
    \centering
    \begin{tikzpicture}
        \matrix[matrix of nodes,
                nodes={inner sep=0, outer sep=0},
                column sep=5mm, row sep=5mm] {
            \node{}; &
            \node{\textbf{Ground Truth}}; &
            \node{\textbf{Prediction}}; &
            \node{\textbf{Error Map}}; \\
            \node{
                \begin{tikzpicture}
                    \node[rotate=90, anchor=center]{\textbf{FNO}};
                \end{tikzpicture}
            }; &
            \node{\includegraphics[width=3cm]{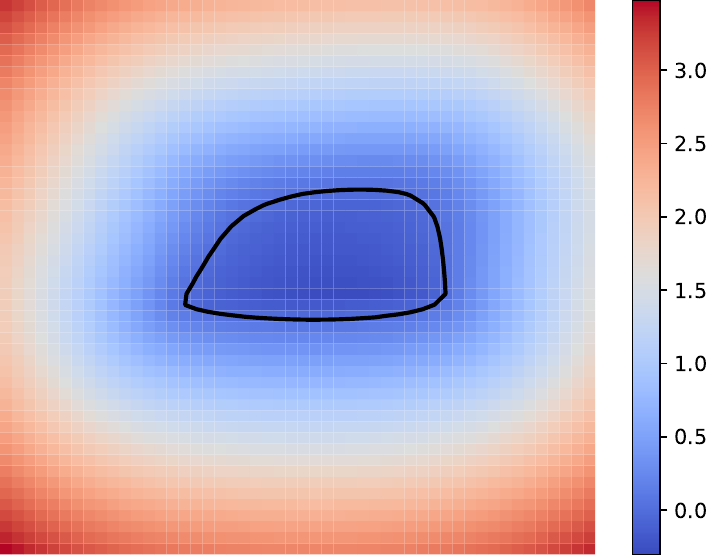}}; &
            \node{\includegraphics[width=3cm]{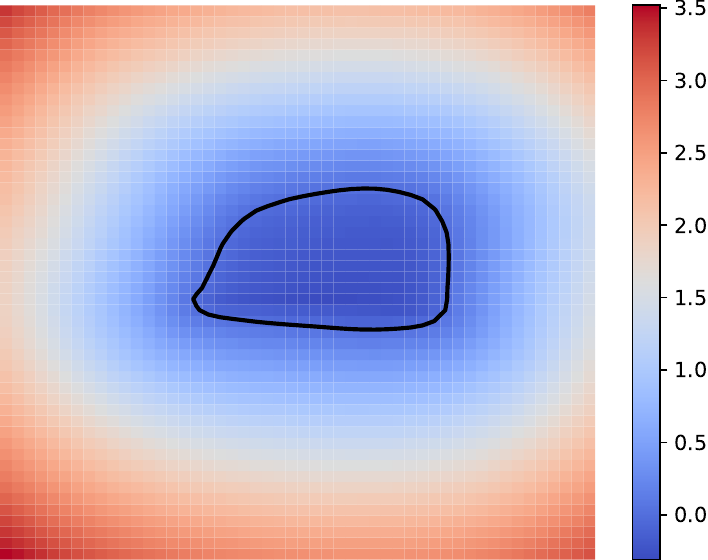}}; &
            \node{\includegraphics[width=3cm]{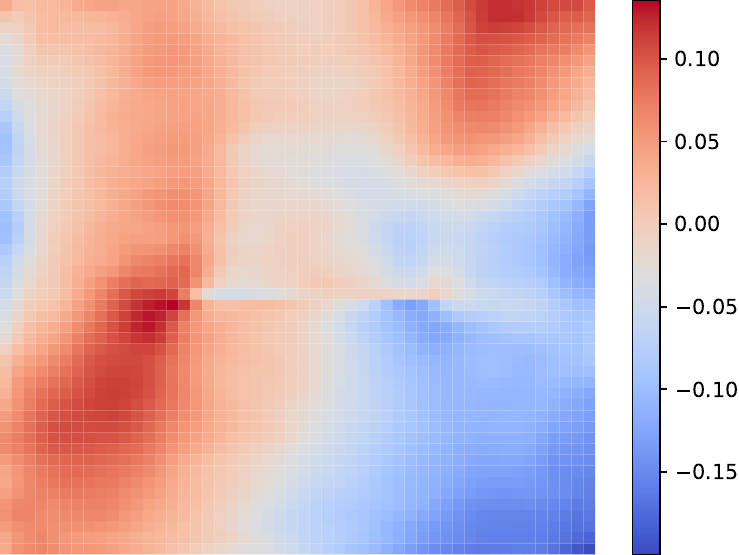}}; \\
            \node{
                \begin{tikzpicture}
                    \node[rotate=90, anchor=center]{\textbf{TNO}};
                \end{tikzpicture}
            }; &
            \node{\includegraphics[width=3cm]{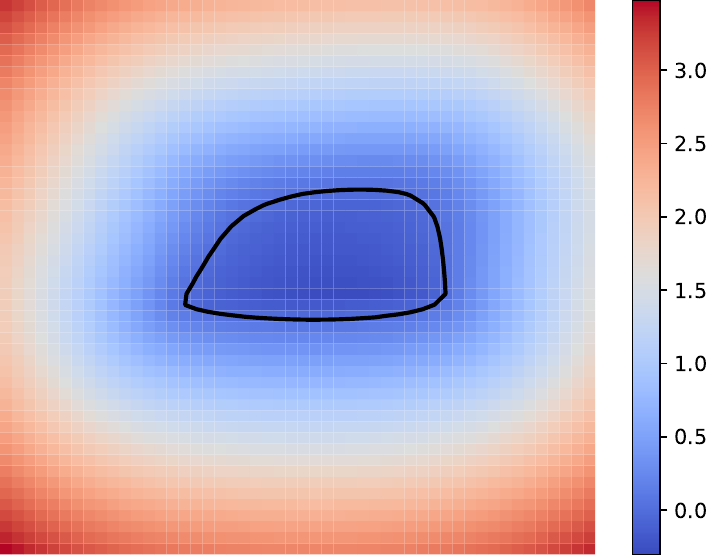}}; &
            \node{\includegraphics[width=3cm]{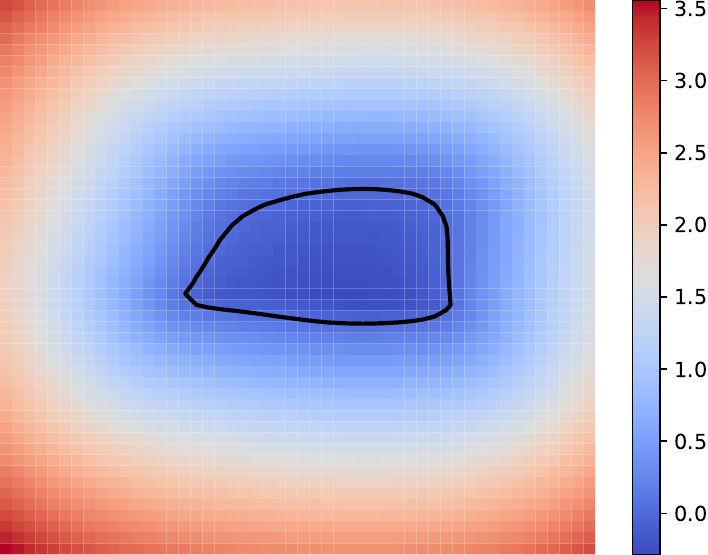}}; &
            \node{\includegraphics[width=3cm]{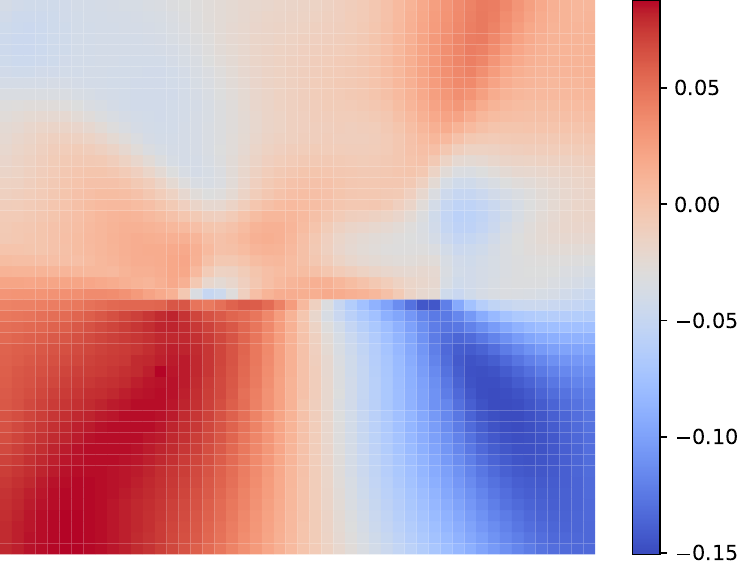}}; \\
        };
    \end{tikzpicture}
    \caption{Velocity-Dependent}
    \label{fig:10}
\end{figure}

\subparagraph{Velocity-Dependent Obstacles.}
In this setup, the obstacle size increases as a function of velocity, simulating environments where faster motion demands greater clearance.
For each sample, the obstacle radius grows with velocity according to either an exponential or logarithmic function, with both the function type and parameters randomly sampled.
An illustration is shown in Fig.\ref{fig:7}.
This setup introduces velocity-dependence in the value function.
We train on 300 and test on 50 such $(V(t=0, x), V_\infty(x))$ pairs.
Examples of predictions are provided in Fig.~\ref{fig:10}.

\begin{figure}[h]
    \centering
    \includegraphics[width=0.8\textwidth]{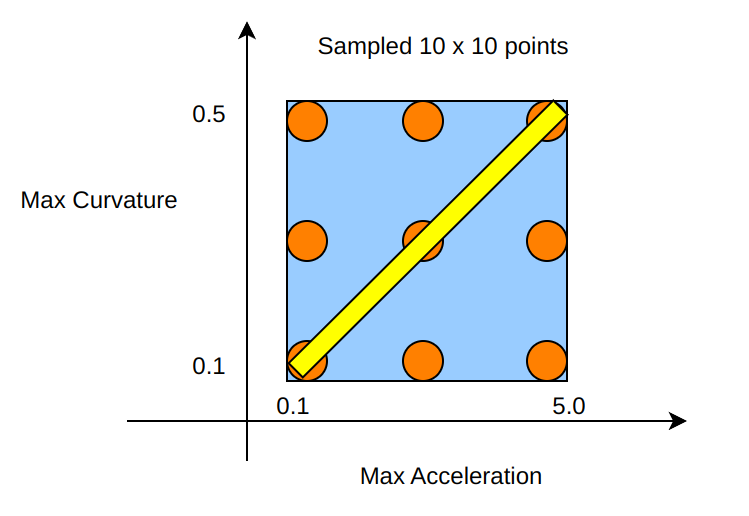}
    \caption{Sampling and testing strategy across varying system hyperparameters}
    \label{fig:3}
\end{figure}

\begin{figure}[h]
    \centering
    \begin{tikzpicture}[scale=0.85]
        \matrix[matrix of nodes,
                nodes={inner sep=0, outer sep=0},
                column sep=3mm, row sep=5mm] {
            \node{
                \begin{tikzpicture}
                    \node[rotate=90, anchor=center]{\textbf{Ground Truth}};
                \end{tikzpicture}
            }; &
            \node{\includegraphics[width=2.8cm]{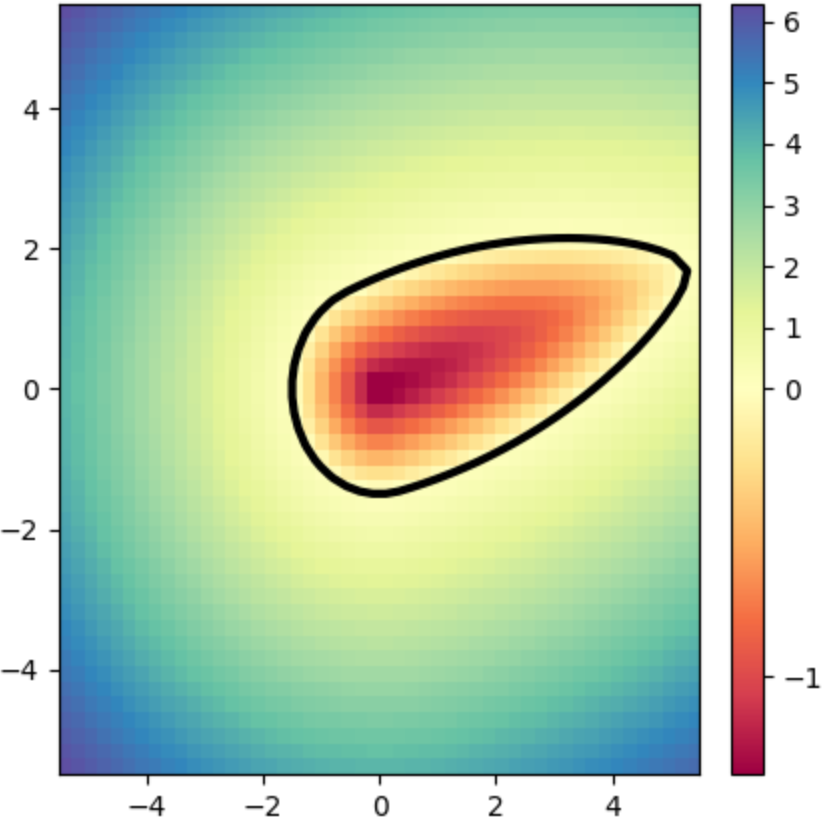}}; &
            \node{\includegraphics[width=2.8cm]{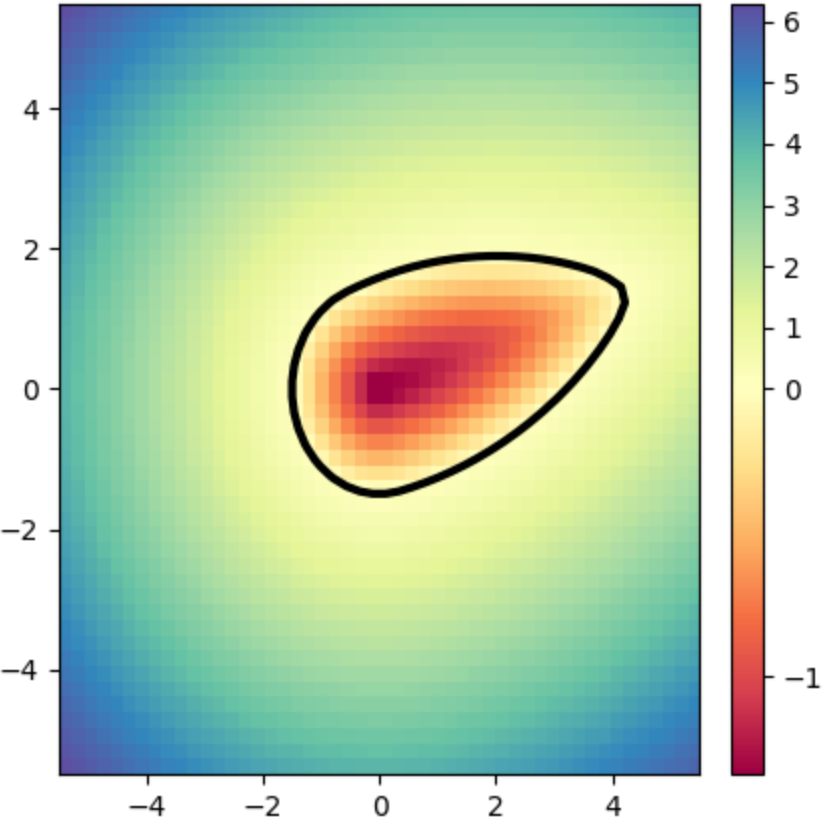}}; &
            \node{\includegraphics[width=2.8cm]{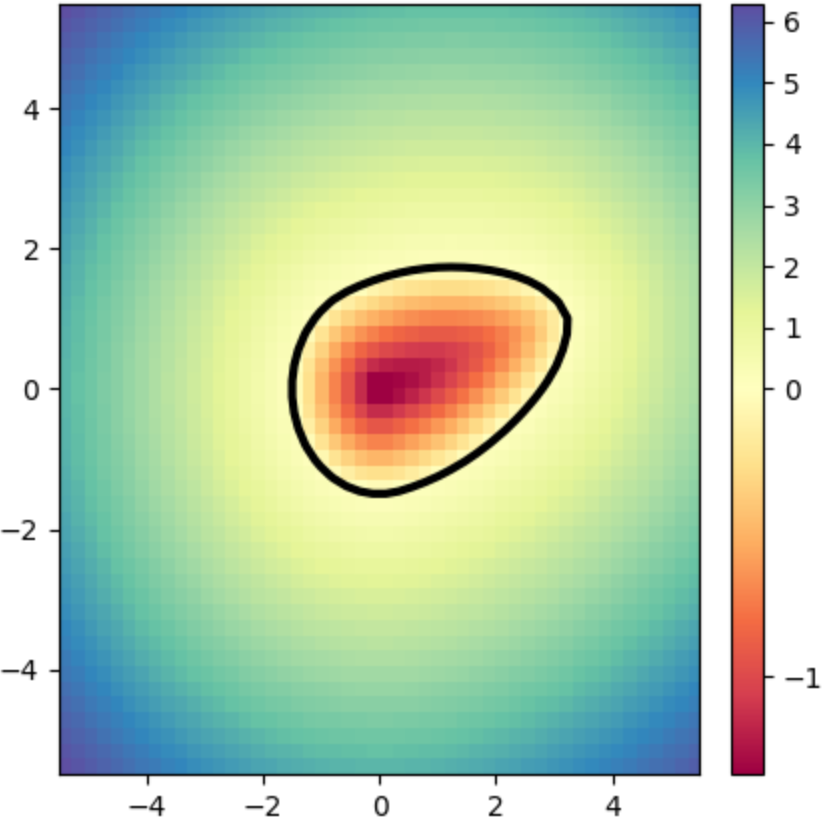}}; &
            \node{\includegraphics[width=2.8cm]{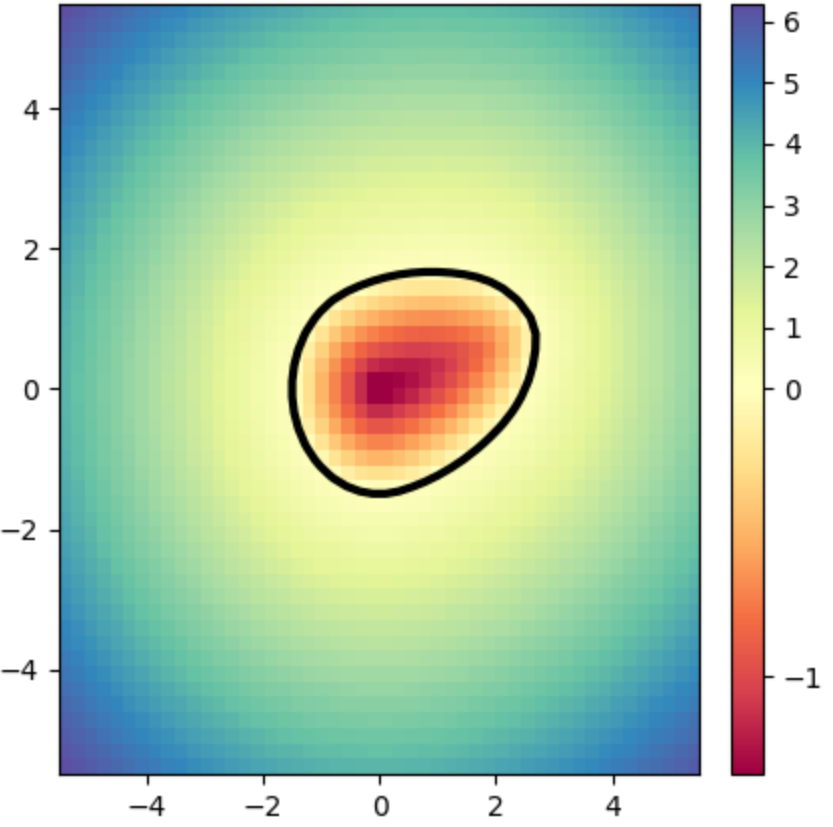}}; &
            \node{\includegraphics[width=2.8cm]{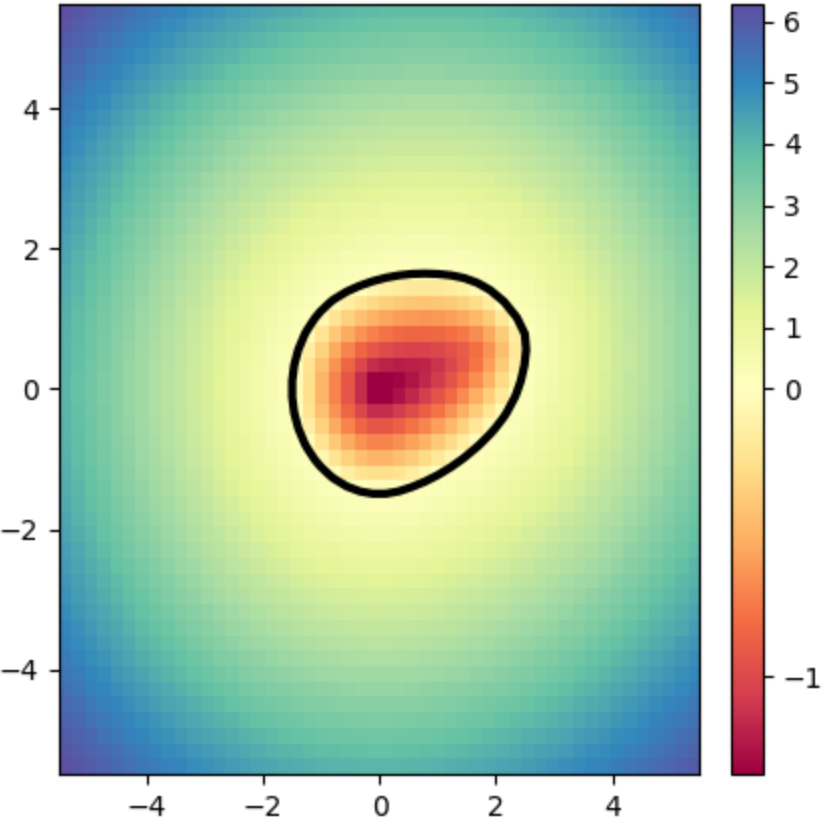}}; \\
            \node{
                \begin{tikzpicture}
                    \node[rotate=90, anchor=center]{\textbf{Prediction}};
                \end{tikzpicture}
            }; &
            \node{\includegraphics[width=2.8cm]{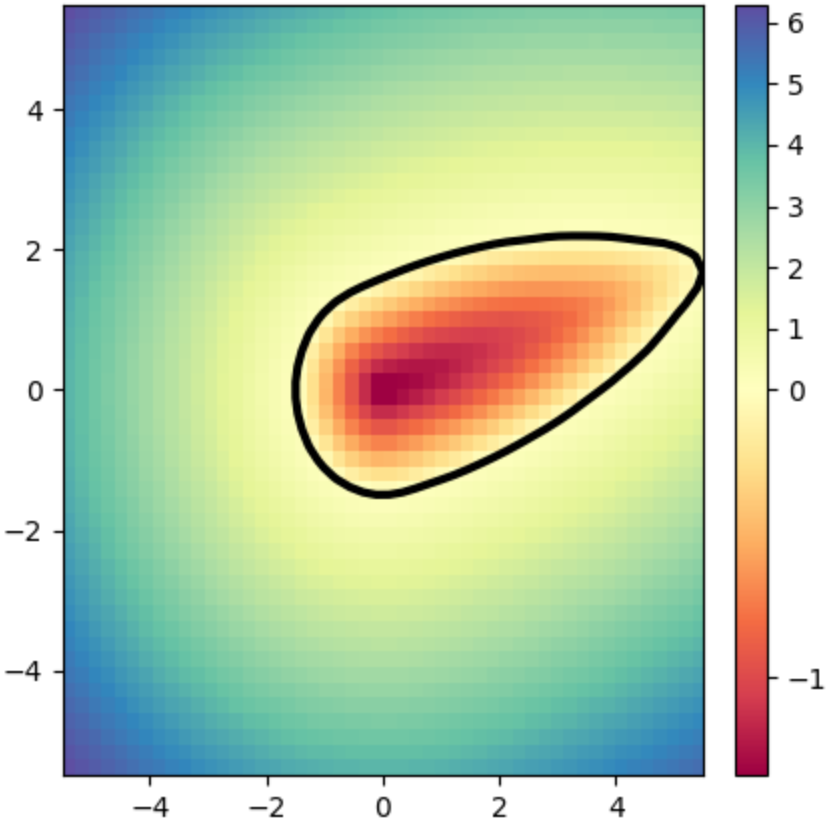}}; &
            \node{\includegraphics[width=2.8cm]{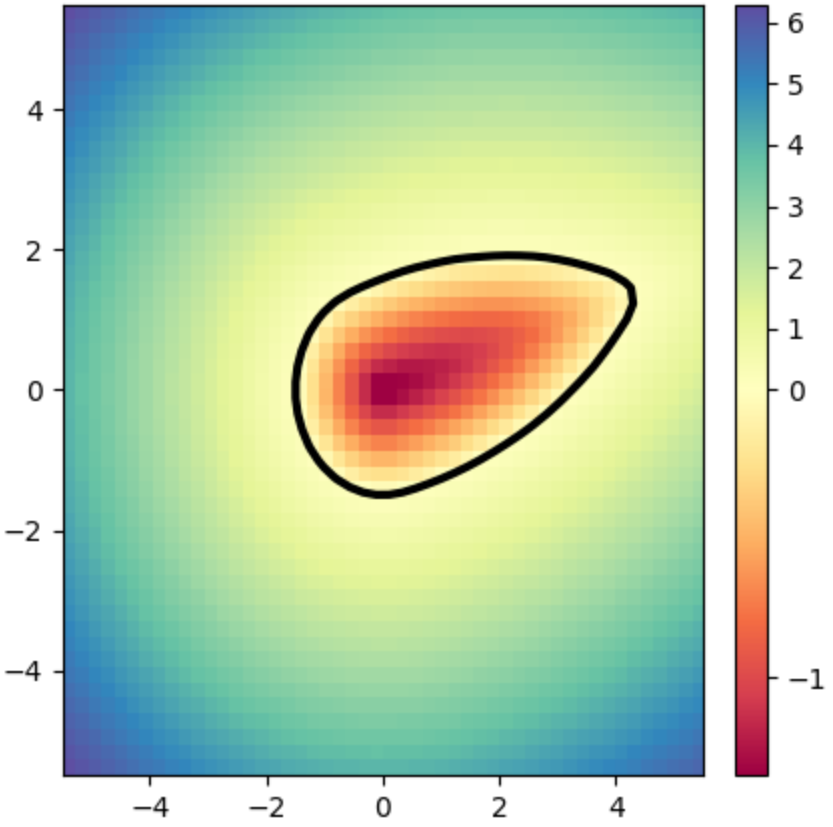}}; &
            \node{\includegraphics[width=2.8cm]{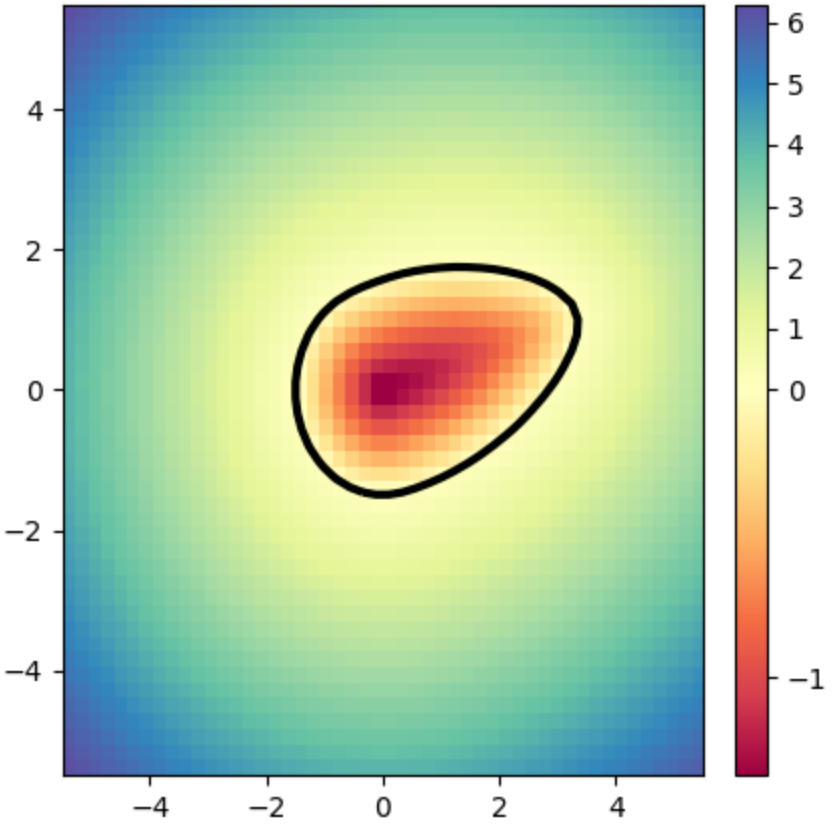}}; &
            \node{\includegraphics[width=2.8cm]{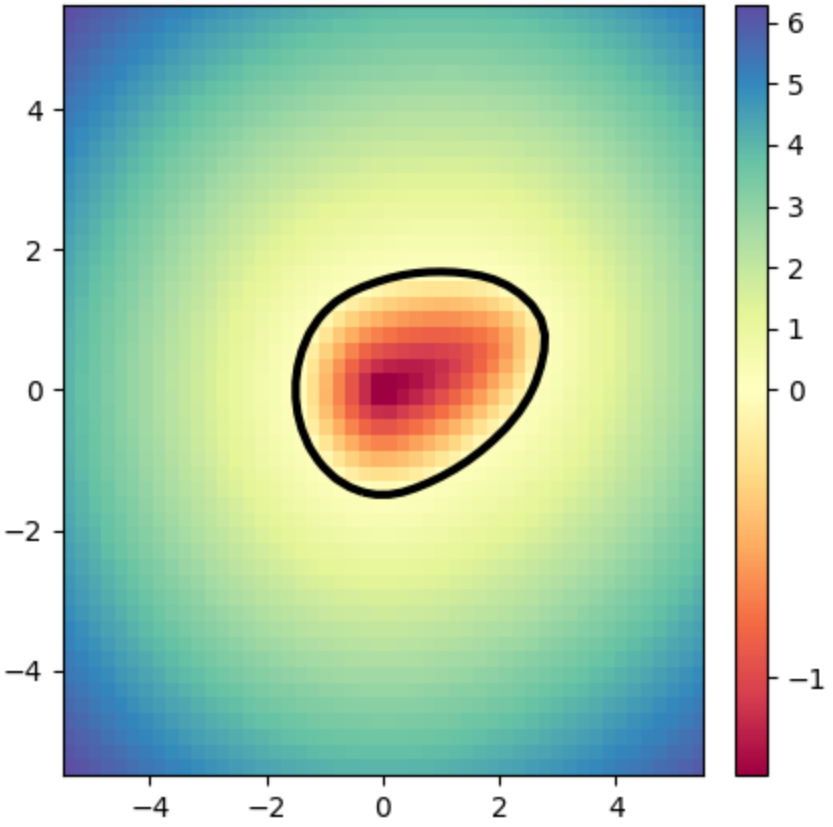}}; &
            \node{\includegraphics[width=2.8cm]{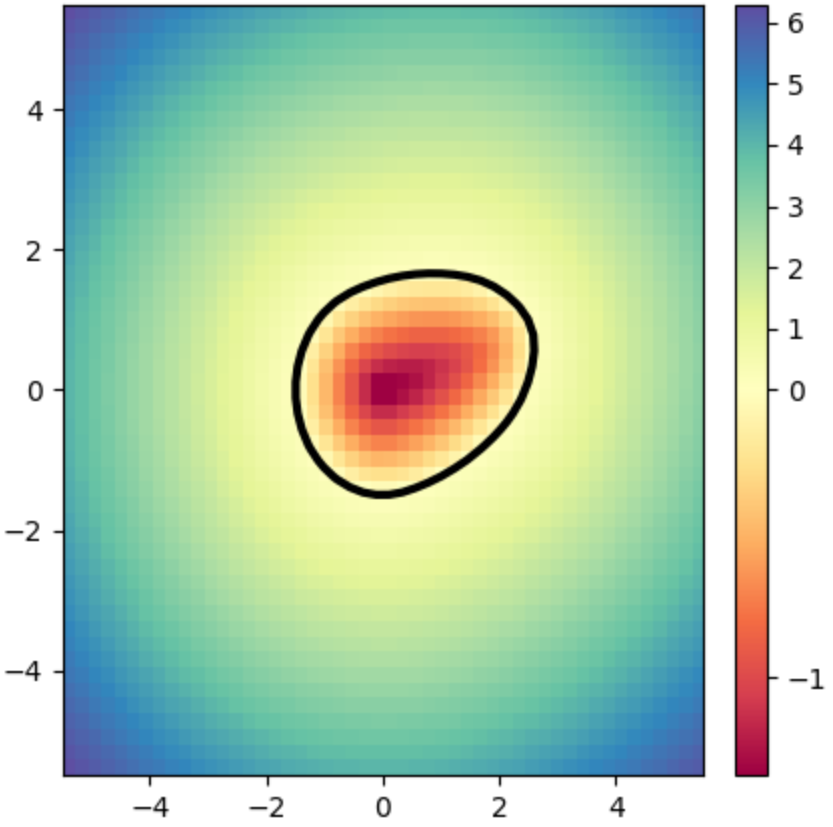}}; \\
        };
    \end{tikzpicture}
    \caption{FNO predictions along the diagonal of the hyperparameter space}
    \label{fig:4}
\end{figure}

\subparagraph{Parametric Inputs.}
To assess generalization over continuous hyperparameter spaces, we vary two control limits: maximum acceleration and maximum curvature.
The training set consists of $10 \times 10 = 100$ combinations from uniformly sampled values along each axis.
We evaluate on 50 test samples randomly drawn with a uniform distribution along the diagonal of the hyperparameter space.
This setting highlights interpolation ability over hyperparameter inputs.
See Fig.~\ref{fig:3} for the sampling strategy and Fig.~\ref{fig:4} for example predictions.

\section{Discussion}

While we presented results on a uniform grid, the method naturally extends to arbitrary geometries \cite{li2023fourier}.
This means HJRNO can be directly applied to environments with non-square or even non-convex domains—for example, an autonomous agent operating within a warehouse with an irregular floor plan.

One limitation is the treatment of hyperparameters.
When introducing additional hyperparameters, each value must be repeated across the hyperparameter domain, increasing the size of the training dataset proportionally.
This approach could become a computational bottleneck when scaling to high-dimensional hyperparameter spaces.
A future research direction could be to design more efficient approaches that better handle hyperparameters.


\bibliographystyle{plain}
\bibliography{references}

\end{document}